\documentclass[journal]{IEEEtran}
\IEEEoverridecommandlockouts                              

\usepackage{siunitx}[=v2]
\usepackage[table,xcdraw]{xcolor}
\usepackage{booktabs}
\usepackage{svg}
\usepackage{cite}
\usepackage{amsfonts}
\usepackage{multirow}
\usepackage[ruled,vlined]{algorithm2e}

\usepackage{float}

\usepackage{enumitem}
\usepackage{mathtools}
\usepackage{amssymb}
\usepackage{epsfig}
\usepackage{epstopdf}
\usepackage{amsmath,hyperref}
\usepackage{bbm}
\SetKwInput{KwIn}{Input}
\SetKwInput{KwInit}{Initialization}
\usepackage{graphicx}
\usepackage{booktabs}
\usepackage{subcaption}

\usepackage[ruled,vlined]{algorithm2e} 

\usepackage[noend]{algpseudocode} 
\usepackage{xcolor}

\makeatletter
\newcommand*{\rom}[1]{\expandafter\@slowromancap\romannumeral #1@}
\makeatother
\usepackage{array}
\SetAlCapFnt{\footnotesize}
\SetAlCapNameFnt{\footnotesize}
\newcommand{\actProceed}{a^{\mathrm{go}}}
\newcommand{\actLeft}{a^{\mathrm{L}}}
\newcommand{\actRight}{a^{\mathrm{R}}}
\newcommand{\actTurn}{a^{\mathrm{U}}}
\usepackage{xcolor}

\usepackage[table,svgnames]{xcolor}
\definecolor{Mahogany}{RGB}{192,64,0}
\definecolor{White}{RGB}{255,255,255}
\definecolor{Bittersweet}{RGB}{254,111,94}

\usepackage[dvipsnames]{xcolor} 

\title{\LARGE \bf
RoboAtlas: Contextual Active SLAM}

\author{Author Names Omitted for Anonymous Review. Paper-ID X.}

\author{Alexander Schperberg$^{1*}$, Shivam K. Panda$^{2*}$, Abraham P. Vinod$^{1}$, Bhagawan Rokaha$^{3}$ \\ M. K. Jawed$^{2}$ , Stefano Di Cairano$^{1}$
}

\begin{document}
\maketitle
\thispagestyle{empty}
\pagestyle{empty}

\begingroup
\renewcommand\thefootnote{} 
\footnotetext{\textsuperscript{*}Denotes equal contribution.}
\endgroup

\footnotetext[1]{A. Schperberg, A. P. Vinod, S. Di Cairano are with Mitsubishi Electric Research Laboratories, Cambridge, MA 02139, USA (email: schperberg@merl.com, abraham.p.vinod@ieee.org, dicairano@ieee.org).}

\footnotetext[2]{S. K. Panda and M. K. Jawed are with the Department of Mechanical and Aerospace Engineering, University of California, Los Angeles, Los Angeles, CA 90095, USA (email: shivamkp@g.ucla.edu, khalidjm@seas.ucla.edu).}

\footnotetext[3]{B. Rokaha is with Mitsubishi Electric Corporation, Japan (email: Rokaha.Bhagawan@df.MitsubishiElectric.co.jp).}

\begin{abstract}
We present \emph{RoboAtlas}, a contextual Active SLAM framework that adaptively balances geometric exploration and semantic reasoning using a scalable 3D semantic mapping system, \emph{OpenRoboVox}. \emph{RoboAtlas} integrates frontier exploration, global semantic-map reasoning, and egocentric VLM-based reasoning through a contextual multi-armed bandit that transitions from exploration to semantically guided navigation as scene understanding improves. We evaluate the system in simulation and on a Unitree Go2 robot in large-scale real-world environments exceeding \SI{1800}{\square\meter} with $\sim$30k mapped semantic instances, achieving a 100\% task success rate. On the GOAT-Bench ``Val Unseen'' benchmark, \emph{RoboAtlas} achieves state-of-the-art performance with highest reported success rate (SR) of 90.6\%, using GPT-4o, improving over the strongest prior baseline by 17.8 percentage points in SR. Using the much smaller Qwen2.5-VL-7B model, it still achieves 88.8\% SR, outperforming all baselines using GPT-4o in SR, and revealing the importance of the information gained by our semantic mapping framework over simply replacing the underlying foundation model. The results demonstrate that grounding foundation models with large-scale 3D semantic maps enables robust and efficient contextual Active SLAM.
\end{abstract}

\section{Introduction}
Active Simultaneous Localization and Mapping (Active SLAM) unifies perception and planning to enable a robot to autonomously navigate, without human intervention, in unknown or unexplored environments. The central objective is to select motions that jointly improve map completeness and accuracy while maintaining a reliable estimate of the robot's pose \cite{Placed2023ActiveSLAM, Schperberg2021Saber}. To this end, Active SLAM typically optimizes trajectories for expected information gain, trading off exploration of unobserved regions against revisiting areas for satisfying practical constraints on time, path length, energy, and safety/traversability. 

\begin{figure}[htbp]
    \centering
    \includegraphics[width=0.99\columnwidth]{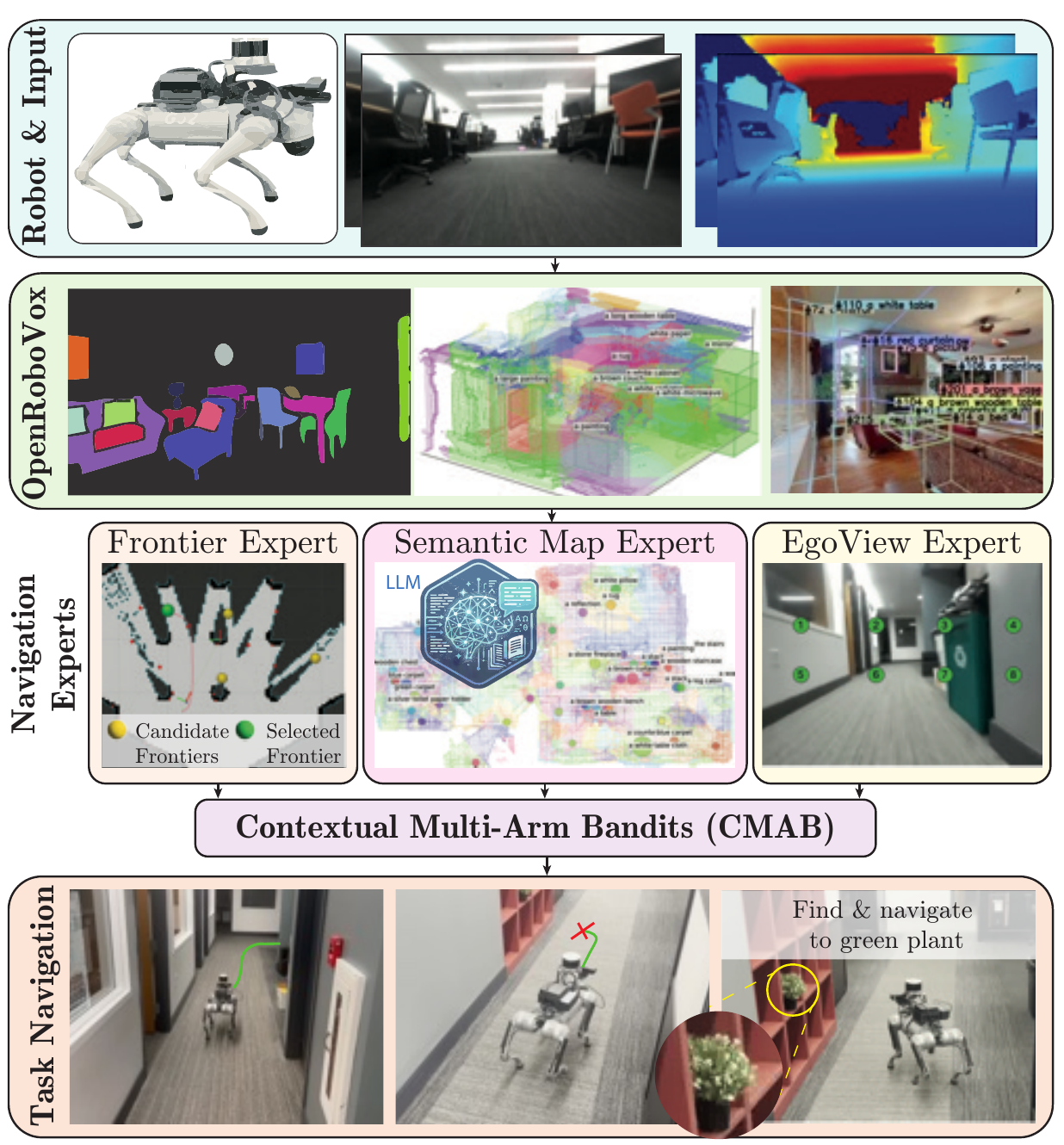}
    \captionsetup{font=small}
    \caption{\textbf{\textit{RoboAtlas}}.
\textit{RoboAtlas} combines frontier exploration, semantic map reasoning, and egocentric VLM reasoning within a contextual multi-armed bandit framework. It receives the environment state through our real-time 3D semantic mapping framework, called \textit{OpenRoboVox}. The system dynamically switches between geometric exploration and semantic navigation as map understanding improves.}
    \label{fig:fig1}
\end{figure}

Existing navigation and exploration pipelines often remain tied to low-fidelity geometric representations such as occupancy grids, and heuristic objectives based on closest-distance or hand-crafted constraints \cite{tokekar2018, Basilico2024}. Such abstractions are often sufficient for coverage tasks, but they limit performance in settings that demand high-level reasoning, including semantic search, inventory tracking, and goal-directed exploration for specific objects or places. In these scenarios, spatial geometry alone is insufficient, as effective decision-making depends on semantic priors about scene structure and object co-occurrence (e.g., forks are likely found in a kitchen) as well as contextual cues that allow extrapolation beyond the current field of view. After all, humans navigate unfamiliar spaces by combining partial observation with rich internal knowledge and language-conditioned expectations. Thus, capturing and operating on this form of contextual reasoning remains challenging within current Active SLAM formulations.

Motivated by this gap, recent works have begun to integrate foundation Large Vision/Language Models (LLM/VLM) into Active SLAM frameworks, which we refer to as \textit{contextual} Active SLAM. Recent approaches are leveraging foundation models in a zero-shot manner to infer out-of-view scene content, enrich semantic maps, and support complex goal specifications without task-specific training or fine-tuning \cite{chen2023train,gadre2022cow,yokoyama2024VFLM}. While attractive for their adaptability, zero-shot strategies may overlook platform-dependent constraints such as localization reliability, map fidelity, and local planning feasibility, and can be suboptimal at different phases of exploration. 

When the map is not well known, pure frontier-based exploration approaches \cite{Karumanchi2025energyConstrained,ko2003practical_frontier, wurm2008coordinated_frontier, benkrid2019multirobot_frontier, Basilico2024, burgard2005coordinated_frontier, yamauchi1999decentralized_frontier} facilitate exploration for map generation, enable optimal path planning at future time steps, and capture sufficient environmental semantics. Once the map has been sufficiently explored, more exploitative behaviors \cite{yokoyama2024VFLM,zhang2025nanollm} enable focusing on areas that are most likely to satisfy the user directive. To this end, we propose a context-adaptive approach, called \textit{RoboAtlas} as shown in Fig. \ref{fig:fig1}, that employs a ``mixture-of-experts" framework. The experts include a frontier-based exploration module, a semantic map module and an egocentric VLM module, the latter two leveraging foundation models to reason over global and local context, respectively. Each expert outputs a candidate goal location for the robot to navigate towards. Taking into account these experts in aggregate, a contextual multi-armed bandit selects the next goal position conditioned on the robot and environment state. In practice we observe that this formulation enables coarse exploration in early deployments, followed by a progressively more targeted, semantically informed navigation as the map and contextual evidence improve. 

Overall, we present the following \textbf{contributions}: 
\begin{enumerate}

\item \textbf{Real-time voxel-based semantic mapping on resource-constrained robots.}
We enable large-scale, real-time voxel-based semantic mapping by combining memory-efficient TSDF (i.e., Truncated Signed Distance Field) optimization, parallel dynamic-instance pruning and updates, and distance-prioritized historical refresh. Our system, \textit{OpenRoboVox}, supports efficient semantic map construction suitable for deployment on edge robotic platforms.

\item \textbf{Foundation model–driven semantic reasoning across global and local context.}
We introduce two complementary experts powered by a shared foundation model: (i) a \textit{semantic map expert} that reasons over a scene dictionary derived from a global 3D map, and (ii) an \textit{egocentric VLM expert} that reasons over the robot’s immediate observations. Together, these modules enable language-guided, semantics-aware goal prediction by integrating global map context with local visual cues.

\item \textbf{Context-adaptive fusion with contextual bandits.}
We employ a contextual multi-armed bandit to dynamically select among frontier-based exploration, a semantic map expert, and an egocentric VLM expert. This yields an adaptive behavior that transitions from efficient coverage in early exploration to targeted semantics-driven search as evidence accumulates.
\end{enumerate}

\subsection{Paper organization}
The paper is organized as follows. Then we provide the related works in Sec. \ref{related_works}. We then provide the overall problem statement with subsections dedicated to a brief overview of the various components of our system, including our semantic mapping framework, \textit{OpenRoboVox}, and the expert algorithms employed by the framework in Sec. \ref{methods:main_sec}. 

More detailed formulation and implementation details of all components are provided in Sec. \ref{sec:implementation}, decomposed into a deeper technical discussion of the bandit algorithm in Sec. \ref{sec:cmab}, \textit{OpenRoboVox} in Sec. \ref{methods:openRoboVox-implementation}, along with its subcomponents such as the Scene Dictionary in Sec. \ref{methods:scene_dictionary}, and details on GPU optimization, and asynchronous parallel architecture for real-time deployment in Sec. \ref{methods:gpu_optimization}, and Sec. \ref{methods:parallel_architecture} respectively. Further abstractions and data representation provided by \textit{OpenRoboVox} to reduce computation load for quick object and semantic search are given in Sec. \ref{methods:2d_pillar}. Implementation details of each expert algorithm are also given: the frontier expert in Sec. \ref{methods:frontier_generation}, the semantic map expert in Sec. \ref{methods:semantic_reasoning}, and the egocentric VLM expert in Sec. \ref{methods:vlm_goal_selection}. Details on the simulation and hardware setups are provided in Sec. \ref{simulation_hardware_setup}. Experimental validation results provided in both simulation and hardware are in Sec. \ref{sec:experimental_validation}. We include both ablation tests as well as extensive baseline comparisons using the Isaac Sim and photo-realistic Habitat simulator environments. Limitations and conclusions of our framework are given in Sec. \ref{sec:limitations} and Sec. \ref{sec:conclusion} respectively.

\section{Related Work}
\label{related_works}
Recent advances in embodied navigation increasingly integrate vision-language models, semantic mapping, and reasoning mechanisms to enable open-vocabulary exploration and object-goal navigation. In this section, we review prior work in three areas most relevant to \textit{RoboAtlas}: (1) zero-shot vision-language and LLM-based navigation, (2) open-vocabulary semantic mapping and scene representations, and (3) learning-based and/or fine-tuning models for semantic exploration and reasoning strategies.

\subsection{Zero-Shot Vision-Language and LLM-based Planning for Navigation}
Recent work has explored object-goal navigation using large Vision-Language Models (VLMs), enabling agents to ground language instructions in visual observations without task-specific training \cite{Dorbala_2024, zhou2023escexplorationsoftcommonsense, huang2023visuallanguagemapsrobot, shah2022lmnavroboticnavigationlarge, huang2022languagemodelszeroshotplanners,huang2022innermonologueembodiedreasoning}. VLFM introduces Vision-Language Frontier Maps, which integrate a VLM-based (BLIP2) similarity scores into a language-conditioned value map that prioritizes frontiers during exploration \cite{yokoyama2024VFLM}. Building on this idea, ASCENT extends zero-shot object-goal navigation to multi-floor environments using a hierarchical floor abstraction and an LLM-driven coarse-to-fine frontier reasoning module \cite{Gong2026}. DyNaVLM improves upon previous zero-shot works by introducing a flexible approach to navigation by allowing agents to freely select navigation targets through visual-language reasoning, instead of relying on fixed motion primitives \cite{ji2025dynavlmzeroshotvisionlanguagenavigation}. Additionally, they include a collaborative graph memory, which is a self-refining graph that encodes spatial relationships and object locations in a topological map, enabling knowledge sharing across agents for improved decision making. However, these approaches primarily rely on image-based perception and topological scene representation that do not maintain a globally consistent metric map of the environment. As a result, semantic reasoning and navigation are often decoupled from geometric localization and mapping. In contrast, \textit{RoboAtlas} integrates vision-language reasoning directly with a real-time Active SLAM pipeline and an instance-level 3D semantic map, enabling language-conditioned decision making while preserving metric consistency required for reliable navigation. 

Beyond perception-driven navigation, another line of work explores the use of LLMs as high-level planners for navigation and decision making. WTRP-Searcher formulates object-goal navigation as a Weighted Traveling Repairman Problem, optimizing the order in which candidate viewpoints are visited based on semantic relevance and travel cost \cite{liu2025handleobjectnavigationweighted}. Such optimization-based planning methods contrast with learned navigation policies and offer improved flexibility across environments. LLMs have also been incorporated as high-level planners for navigation. OrionNav combines an LLM planner with online semantic mapping and hierarchical scene graphs to generate context-aware navigation plans that adapt as new information becomes available \cite{devarakonda2024orionnavonlineplanningrobot}. Other works use LLMs in more limited roles. For instance, 3P-LLM combines LLM preferences with feasibility scores to guide navigation decisions \cite{latif20243pllmprobabilisticpathplanning}. In a different problem setting, NAMO-LLM leverages LLM guidance to bias sampling-based planners for navigation among movable obstacles in 2D cluttered environments \cite{zhang2025nanollm}. Huang \textit{et al.}~\cite{MSGNav} introduce MSGNav, a zero-shot navigation framework built on a multi-modal 3D scene graph that preserves visual evidence within relational representations and supports open-vocabulary reasoning. 

While these methods demonstrate the potential for LLMs for high-level decision making, they typically operate on simplified 2D representations or rely on predefined planning abstractions. \textit{RoboAtlas} differs by embedding foundation model reasoning directly within a 3D instance-level semantic map and coupling it with a contextual decision policy. In particular, our framework uses a contextual multi-armed bandit to adaptively fuse geometric exploration, semantic map reasoning, and egocentric VLM cues, enabling the system to transition from broad exploration to targeted semantic search as contextual evidence accumulates.   

\subsection{Open-Vocabulary Semantic Mapping and Scene Representation}

Effective navigation in large environments often requires persistent semantic representations of the scene.
Recent work has explored open-vocabulary mapping and scene representations that combine geometric mapping with semantic understanding \cite{mccormac2016semanticfusiondense3dsemantic, rünz2018maskfusionrealtimerecognitiontracking, mccormac2018fusionvolumetricobjectlevelslam, narita2019panopticfusiononlinevolumetricsemantic, qian2022pocdprobabilisticobjectlevelchange, qian2023povslamprobabilisticobjectawarevariational}.
ConceptGraphs~\cite{gu2024conceptgraphs} constructs open-vocabulary 3D scene graphs by lifting outputs from 2D foundation models into an instance-level map, where detected objects become nodes connected by spatial relationships, and associates observations across frames using IoU-based matching.
OpenVox~\cite{deng2025openvoxrealtimeinstancelevelopenvocabulary} targets the noise sensitivity of such threshold-based association by replacing the point-cloud scene graph with a probabilistic instance voxel grid, in which each voxel maintains a distribution over instance IDs and semantics are stored in a separate caption-embedding codebook. It casts cross-frame fusion as two probabilistic subtasks, instance association and live map evolution, which improves robustness to segmentation noise and supports incremental real-time updates.
OneMap~\cite{busch2024onemap} proposes reusable open-vocabulary feature maps for multi-object navigation, allowing agents to reuse information gathered during previous searches.

These systems typically rely on open-vocabulary perception models for object detection and segmentation.
GroundingDINO enables language-conditioned object detection \cite{liu2023groundingdino}, while YOLO-World focuses on real-time open-vocabulary detection for robotics applications \cite{cheng2024yoloworld}.
Several works rely on CLIP-based similarity to align language queries with visual observations \cite{radford2021clip}, or VLMs such as BLIP-2 \cite{li2023blip2}.
For segmentation, promptable models such as SAM \cite{kirillov2023sam} and lightweight variants like MobileSAM \cite{zhang2023mobilesam} enable open-vocabulary object mask generation.
FC-CLIP further extends these ideas to open-vocabulary semantic segmentation using frozen CLIP representations \cite{yu2023fcclip}.

While these approaches provide powerful representations for open-vocabulary semantic mapping, many are primarily evaluated in simulation, offline settings, or small maps, and their scalability for continuous real-time deployment on physical robots in larger maps remains less explored. In particular, OpenVox demonstrates strong instance-level semantic mapping capabilities, but its original implementation assumes idealized conditions such as perfect localization and unconstrained GPU resources. In contrast, \textit{RoboAtlas} builds upon OpenVox and introduces \textit{OpenRoboVox}, a re-engineered framework designed for real-time deployment on physical robots. Our modifications include memory-efficient TSDF management, asynchronous semantic processing, a reduced-order 2D pillar map representation for fast inference, and a scene-dictionary representation that enables scalable reasoning over large environments while maintaining stable real-time mapping performance.

\subsection{Learning-based Semantic Exploration and Reasoning for Embodied Navigation}

Another line of work integrates semantic reasoning and learning-based strategies into embodied exploration \cite{das2017embodiedquestionanswering, li2024behavior1khumancenteredembodiedai, ren2024exploreconfidentefficientexploration, shridhar2020alfredbenchmarkinterpretinggrounded, szot2022habitat20traininghome}. For example, \cite{zhu2025mtu} proposes Move-to-Understand (MTU3D), which unifies visual grounding and exploration through large-scale vision-language pretraining, and learns spatial memory directly from RGB-D observations to jointly optimize object grounding and frontier-based exploration.

However, these approaches primarily rely on large-scale pretraining or structured scene representations for reasoning, which can introduce significant computational overhead and limit direct deployment on physical robots. In contrast, \textit{RoboAtlas} improves the exploration decision layer within an Active SLAM pipeline and leverages \textit{OpenRoboVox} for hardware-efficient instance-level semantic mapping. This enables real-time operation on physical robots while still incorporating semantic cues, allowing exploration strategies to adapt to the evolving scene context without requiring large-scale training or expensive scene-graph reasoning.

Several recent works also investigate memory-augmented exploration. For example, 3D-Mem \cite{yang20253dmem} is a 3D scene memory framework for embodied AI agents that uses multi-view “memory snapshots” and frontier exploration to help robots efficiently explore, remember, and reason about complex indoor environments over long periods. ReEXplore \cite{zhang2025reexplore} improves frontier selection through retrospective experience replay, LMEE~\cite{wang2026explore} studies long-term episodic memory for multi-goal navigation and memory-based question answering, and HIMM~\cite{li2026himmhumaninspiredlongtermmemory} introduces human-inspired episodic and semantic memory for exploration reasoning. While these methods enhance exploration through memory retrieval and experience-based reasoning, they generally operate within fixed exploration policies or require additional training, whereas \textit{RoboAtlas} does not have training requirements due to the bandit approach adjusting the behavior during deployment. 

\section{Contextual Active SLAM}
\label{methods:main_sec}











\begin{algorithm}[t]
\caption{\textit{RoboAtlas} Framework}
\label{alg:roboAtlas}
\KwIn{Language directive $\mathcal{L}_T$\;
expert set $\mathcal{E}=\{E_{\mathrm{Frontier}}, E_{\mathrm{SemanticMap}}, E_{\mathrm{EgoVLM}}\}$\;
goal-selection policy $\pi$ (Sec.~\ref{sec:cmab})}
\KwInit{Bring up SLAM stack, \textit{OpenRoboVox}, and asynchronous perception threads\;
Scene-Dictionary $\mathcal{S}_0 \leftarrow \varnothing$ (Sec. \ref{methods:scene_dictionary})\;
Initialize policy $\pi$; \quad decision epoch $t \leftarrow 0$}
\While{task $\mathcal{L}_T$ not completed}{
  Read latest pose, occupancy grid $\mathbf{M}_t^{\rm grid}$, and Scene-Dictionary $\mathcal{S}_t$\;
  Build contextual state $\mathbf{c}_t$ from $\mathbf{M}_t^{\rm grid}$, $\mathcal{S}_t$, ego pose, and $\mathcal{L}_T$\;
  \ForEach{expert $E_a \in \mathcal{E}$}{
  Propose goal $\mathbf{g}_t^{(a)} \leftarrow E_a(\mathbf{c}_t, \mathcal{S}_t, \mathbf{M}_t^{\rm grid}, \mathcal{L}_T)$\;
}
Collect admissible arms $\mathcal{A}_t \leftarrow \{a \in \mathcal{A} : \mathbf{g}_t^{(a)} \text{ is valid}\}$\;
Select arm $a_t \leftarrow \pi(\mathbf{c}_t, \mathcal{A}_t)$ and set goal $\mathbf{g}_t \leftarrow \mathbf{g}_t^{(a_t)}$\;
  Dispatch $\mathbf{g}_t$ to the navigation stack; execute until arrival or replanning event\;
  Observe reward $r_t$ (Eq.~\ref{eq:reward}) and update $\pi$ with $(\mathbf{c}_t, a_t, r_t)$\;
  $t \leftarrow t + 1$\;
}
\Return semantic map, Scene-Dictionary $\mathcal{S}_t$, task result\;
\end{algorithm}

\begin{figure*}[!t]
    \centering
    
    \begin{subfigure}{\textwidth}
        \centering
        \includegraphics[width=\textwidth]{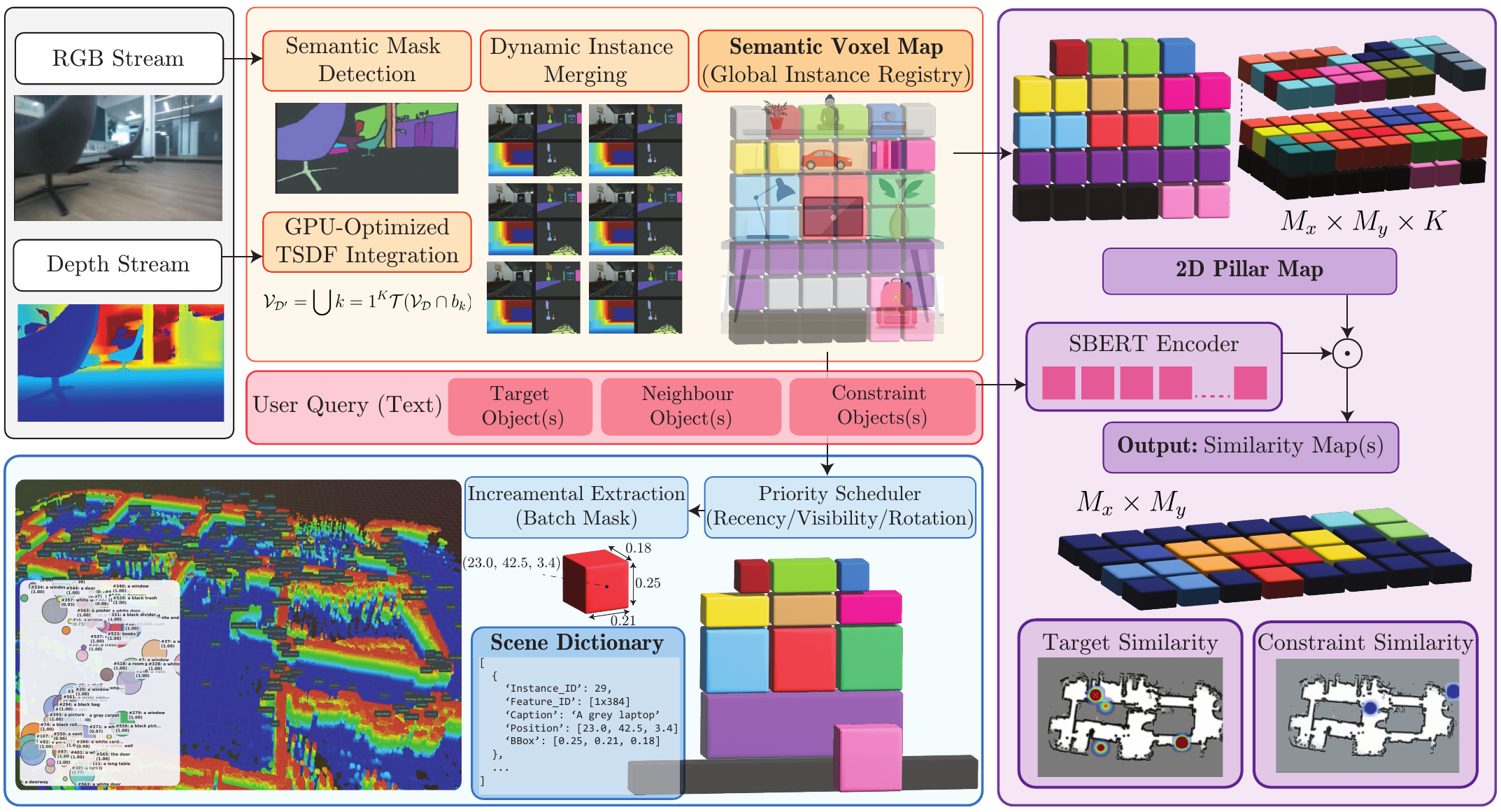}
        \caption{\textit{OpenRoboVox}}
        \label{fig:openrobovox}
    \end{subfigure}
    
    \vspace{1em}
    
    \begin{subfigure}{\textwidth}
        \centering
        \includegraphics[width=\textwidth]{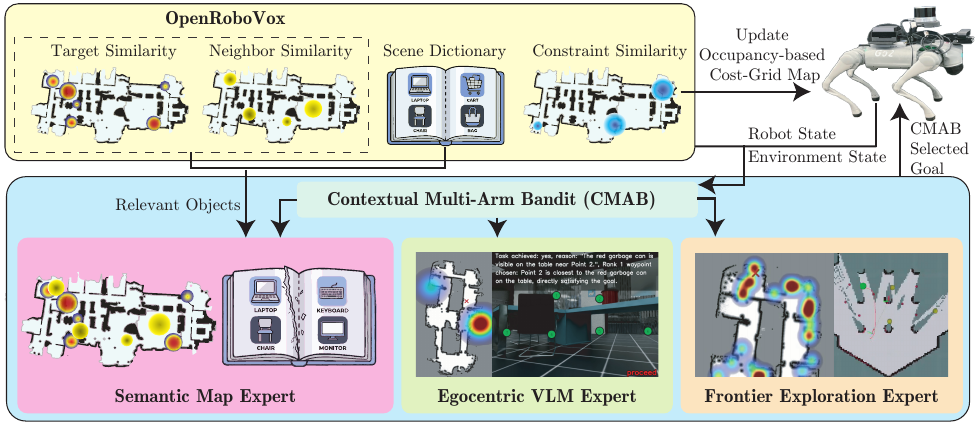}
        \caption{Contextual Multi-Arm Bandits}
        \label{fig:2b-roboatlas}
    \end{subfigure}
    
    \captionsetup{font=small}
    \caption{\textbf{\textit{RoboAtlas} overall framework.}
Top: \textit{OpenRoboVox} performs real-time 3D semantic mapping and scene-dictionary construction from RGB-D observations. Bottom: a contextual multi-armed bandit selects among frontier exploration, semantic map, and egocentric VLM experts to generate navigation goals.}
    \label{fig:robo_atlas}
\end{figure*}

\subsection{Problem Statement}
\label{methods:prelim}

We consider a mobile robot operating in a previously unexplored environment while executing a high-level language-conditioned task. The robot must simultaneously localize itself, construct a semantic representation of the environment, and select navigation goals that maximize the likelihood of task completion.

A fundamental challenge in Contextual Active SLAM is that the utility of different navigation strategies changes throughout deployment. When little is known about the environment, geometric exploration is necessary to acquire map coverage and semantic observations. As environmental understanding improves, however, decisions should increasingly leverage semantic context, object relationships, and task-specific reasoning. Existing approaches typically rely on either fixed exploration policies, fixed semantic reasoning strategies, or semantic representations that are insufficiently rich to support contextual reasoning, making them suboptimal across different stages of exploration.

To address this challenge, we formulate Contextual Active SLAM as an adaptive goal-selection problem. Specifically, we consider a set of expert policies

\begin{equation}
\mathcal{E}
=
\left\{
E_{\mathrm{Frontier}},
E_{\mathrm{SemanticMap}},
E_{\mathrm{EgoVLM}}
\right\},
\end{equation}
where each expert proposes navigation goals using different sources of information. The frontier expert, $E_{\rm Frontier}$, prioritizes geometric information gain, the semantic-map expert, $E_{\rm SemanticMap}$, reasons over a persistent semantic global map, and the ego-centric VLM expert, $E_{\rm EgoVLM}$, reasons directly over the robot's current viewpoint. Rather than relying on a fixed expert or predefined weighting, our objective is to adapt expert selection according to the robot's evolving understanding of the environment.

Formally, we consider a robot capable of performing Active Simultaneous Localization and Mapping (SLAM), enabling autonomous navigation within an environment that may be partially or completely unknown \textit{a priori}. At discrete decision epochs $t=1,2,\ldots,T$, the robot selects a desired goal position $\mathbf{g}_t \in \mathbb{R}^3$. A 3D OctoMap \cite{hornung13auro} generated from LiDAR observations is projected into an occupancy grid $\mathbf{M}_t^{\rm grid}$ for navigation and obstacle avoidance, while localization is provided by the methods in \cite{Macenski2021}.

To extend traditional Active SLAM into \textit{Contextual Active SLAM}, we assume the robot receives a high-level language directive $\mathcal{L}_T$ that remains active until completion. The directive may be exploration-driven (e.g., ``search the area as best as possible''), object-centric (e.g., ``find object $A$ next to object $B$, then object $C$''), or inventory-oriented (e.g., ``keep track of objects $A$--$C$'').

Let $\mathbf{c}_t$ denote the robot's contextual state, including geometric, semantic, and task-relevant information. The objective is to determine a navigation goal according to

\begin{equation}
(\mathcal{L}_t,\mathbf{c}_t)
\;\xrightarrow{\;\;\pi (\mathcal{E})\;\;}
\mathbf{g}_t \in \mathbf{M}_t^{\rm grid},
\label{eq:prompt_to_goal}
\end{equation}
where $\pi$ denotes the contextual goal-selection policy. The key objective is to learn a policy that transitions naturally from exploration-driven behavior, when environmental knowledge is limited, to semantically guided navigation as contextual evidence accumulates.

Overall, the \textit{RoboAtlas} framework consists of two main components: \textit{OpenRoboVox} (Sec.~\ref{methods:openRoboVox-concept} \& \ref{methods:openRoboVox-implementation}), which provides 3D semantic mapping of the surrounding environment, and a goal-selection policy based on contextual bandits (Sec.~\ref{sec:cmab}) that leverages the \textit{OpenRoboVox} outputs. The bandits algorithm negotiates among a mixture-of-expert algorithms, $\mathcal{E}$ (see Sec. \ref{methods:mixture_of_experts}), based on the robot's current state. Fig.~\ref{fig:robo_atlas} illustrates the complete \textit{RoboAtlas} framework, with the overall algorithm of our methods formulated in Algorithm \ref{alg:roboAtlas}.

\subsection{OpenRoboVox}
\label{methods:openRoboVox-concept}

\textit{OpenRoboVox} is our open-vocabulary semantic mapping framework that enables long-horizon semantic navigation on real-world mobile robots. Beyond maintaining a large-scale 3D geometric representation of the environment, \textit{OpenRoboVox} continuously distills raw voxel-level observations into a structured semantic memory called the Scene-Dictionary, denoted as $\mathcal{S}_t$. The Scene-Dictionary provides a compact, queryable representation of objects, their semantic descriptions, and their spatial properties, condensing millions of low-level voxels into a representation suitable for high-level reasoning. This abstraction enables downstream expert policies to efficiently perform semantic retrieval, language-guided navigation, and LLM-based scene reasoning without directly operating on the underlying voxel map. \textit{OpenRoboVox} therefore serves as the semantic world model that underpins the semantic map, VLM, and exploration experts described in the following sections. Implementation details of \textit{OpenRoboVox} are provided in Sec. \ref{methods:openRoboVox-implementation}.

\subsection{Mixture-of-Experts Selection via Bandits}
\label{methods:mixture_of_experts}

A mixture-of-experts policy, using contextual bandits, is employed to select the expert, at the current state, that maximizes rewards such as completing the user-desired objective, map expansion, and reducing backtracking. Further details are provided in Sec. \ref{sec:cmab}. We provide an overview of each expert in Sec. \ref{methods:frontier_selection}, \ref{methods:semantic_map_expert}, \ref{methods:vlm}, with further implementation details in Sec. \ref{sec:implementation}.

\subsubsection{Frontier Exploration Expert}
\label{methods:frontier_selection}

The frontier exploration expert selects navigation goals that expand the robot's knowledge of the environment, while ensuring that exploration remains compatible with the robot's energy constraints. Intuitively, frontier-based exploration seeks regions at the boundary between explored and unexplored space, since observations collected near these boundaries are likely to reveal previously unknown areas. The expert prioritizes goals that are expected to provide high information gain while avoiding decisions that could compromise the robot's ability to safely reach a charging station.

At each replanning time $t$, the expert selects a navigation goal $\mathbf{g}_t$ by evaluating a set of frontier candidates and choosing the one that best balances exploration utility and travel cost under energy-feasibility constraints. The detailed frontier generation procedure and optimization objective are described in Section~\ref{methods:frontier_generation}. The full description of the frontier selection is given in \cite{Karumanchi2025energyConstrained}.

\subsubsection{Semantic Map Expert}
\label{methods:semantic_map_expert}

While the Scene-Dictionary $\mathcal{S}_t$, described in Sec. \ref{methods:openRoboVox-concept}, provides a structured, queryable representation of the environment, its full utility emerges when coupled with the reasoning capabilities of an LLM. Unlike traditional semantic maps that rely on fixed-schema queries (e.g., ``Find object class $C$''), our framework allows the LLM to reason directly over the evolving semantic state of the environment.

Conditioned on the current contents of $\mathcal{S}_t$, the LLM can answer open-vocabulary queries, resolve referential ambiguity, and infer spatial relationships between objects that are not explicitly encoded in the map representation. This enables the semantic map expert to serve as a high-level semantic reasoning module within the bandits framework. In addition to supporting interactive querying, the expert can evaluate candidate object instances and estimate their plausibility as navigation targets by jointly considering semantic evidence and scene context.

The implementation of the semantic reasoning pipeline, including context construction, instance filtering, and confidence scoring, is described in Sec.~\ref{methods:semantic_reasoning}.

\subsubsection{Egocentric VLM Expert}
\label{methods:vlm}

The egocentric VLM expert enables goal selection directly from the robot's current visual observations. Unlike the semantic map expert, which reasons over a persistent world model, the VLM expert operates on the robot's instantaneous first-person view and is therefore particularly useful when target-relevant information is visible but has not yet been incorporated into the semantic map.

At each replanning step $t$, the expert jointly interprets the user query $\mathcal{L}_{t}$, and the current scene observation. Based on these inputs, the VLM estimates the relevance of visible candidate targets and determines whether the robot should continue toward a candidate location or first reorient its viewpoint. This capability allows the system to recover from challenging visual configurations, such as occlusions, poor viewing angles, or situations in which the desired target is not currently visible.

In addition to action and goal selection, the VLM generates textual scene descriptions that summarize the observed environment. These descriptions are accumulated into the scene context $D_{\rm scene}$, which is subsequently provided to the semantic map expert described in Sec.~\ref{methods:semantic_map_expert}, enabling richer semantic reasoning across experts. Implementation details of the Egocentric VLM is provided in Sec.~\ref{methods:vlm_goal_selection}.

\section{Implementation Details}
\label{sec:implementation}

\subsection{Contextual Multi-Arm Bandits}
\label{sec:cmab}

For $\pi$ in Eq. \ref{eq:prompt_to_goal}, we employ Contextual Multi-Arm Bandits (CMAB) \cite{Li_2010}, as formulated by Algorithm \ref{alg:linucb}. In CMAB, $\pi$ selects an arm from a set of arms denoted by $\mathcal{A}={1,\ldots,A}$, where each arm $a\in\mathcal{A}$ is a goal-proposed \textit{information gain expert}, $\mathcal{E}$, that returns a single candidate goal $\mathbf{g}^{(a)}_t\in \mathbf{M}_t^{\rm grid}$. In our setting, $A=3$, where $E_\text{Frontier}$ is the expert based on frontier exploration described fully in \cite{Karumanchi2025energyConstrained} and summarized in Sec. \ref{methods:frontier_generation}; $E_{\text{SemanticMap}}$ denotes the semantic map expert obtained by querying a foundation model using a scene-dictionary as described in Sec. \ref{methods:semantic_reasoning}, derived from a 3D semantic map built using \textit{OpenRoboVox} described in Sec.~\ref{methods:openRoboVox-implementation}; and $E_{\text{EgoVLM}}$ is the expert from an egocentric VLM that uses the current RGB and depth images as input, see Sec.~\ref{methods:vlm_goal_selection}.

Let $\mathcal{A}_t\subseteq\mathcal{A}$ denote subset of arms with valid proposals at epoch $t$. We summarize the robot's current exploration and semantic state by the context vector
$ \mathbf{c}_t = \big[ m_t^{\text{occ}},\ \dot{m}_t^{\text{occ}},\ B_t,\ a_{t-1}\, V_t^{\text{vlm}},\ V_t^{\text{sim}} \big] $
where $m_t^{\rm occ}$ is the grid coverage, $\dot{m}_t^{\rm occ}$ is the change in grid coverage from time step $t$ to $t-1$,  $B_t$ is a backtracking indicator, and $a_{t-1}$ the previous action chosen by CMAB. $V_t^{\text{vlm}}$ is a metric for how confident the egocentric VLM is in its decision, see Sec. \ref{methods:vlm_goal_selection}), and $V_t^{\text{sim}}$ is a similarity-based metric that aggregates target relevance across precomputed text embeddings and the scene dictionary, defined next.

The similarity-based context feature $V_t^{\rm sim}$ aggregates target relevance across detected objects using the similarity score in Eq.~\ref{eq:sim_score}. Let $\mathcal{O}_t$ denote the set of candidate objects observed at epoch $t$, and let $s_o=\mathcal{T}_{\rm sim}(o,\mathcal{Q}_{\mathcal{T}})$ denote the similarity score between object $o$ and the task query $\mathcal{Q}_{\mathcal{T}}$ (see Sec. \ref{methods:similarity} for full definition of the similarity score). Task query $\mathcal{Q}_{\mathcal{T}}$, is derived from the general user-directed query $\mathcal{L}_t$ (e.g., from "Go to the red chair" to "red chair").

We define three similarity thresholds $0 < \tau_1 < \tau_2 < \tau_3 \le 1$ and count the number of observed objects exceeding each,
\begin{equation}
N_k(t) = \sum_{o \in \mathcal{O}_t} \mathbf{1}\!\left(s_o \ge \tau_k\right), \quad k \in \{1,2,3\}.
\end{equation}
The similarity context feature aggregates these counts, weighting higher-confidence tiers more heavily:
\begin{equation}
\label{eq:v_sim}
V_t^{\text{sim}} = \tau_1 N_1(t) + \tau_2 N_2(t) + \tau_3 N_3(t).
\end{equation}
In Eq. \ref{eq:v_sim}, an object with similarity in the top tier contributes $\tau_1 + \tau_2 + \tau_3$ (since it passes all three thresholds), while an object in the lowest tier contributes only $\tau_1$. This produces a soft-count signal that grows with both the number and confidence of target-relevant observations.

We use LinUCB (Linear Upper Confidence Bound) \cite{Li_2010}, which is a contextual bandit algorithm that extends the UCB principle to linear reward models. It assumes each arm’s expected reward is a linear function of context features and selects arms optimistically using an upper confidence bound, thereby balancing exploitation and exploration. At time step $t$, the agent observes a context $\mathbf{c}_t$. Each available arm $a\in\mathcal{A}_t$ is represented by a $d$-dimensional feature vector $\mathbf{x}_{t,a}\in\mathbb{R}^d$ derived from $\mathbf{c}_t$. For online execution, we maintain for each
arm a covariance matrix and response vector,
\(
\mathbf{A}_a \in \mathbb{R}^{d\times d}
\)
and
\(
\mathbf{b}_a \in \mathbb{R}^{d},
\)
which are updated incrementally as new data arrive. We encode each candidate arm \(a\) under context \(\mathbf{c}_t\) using the arm-aware
feature map $\phi(\mathbf{c}_t,a)\in\mathbb{R}^d$.
This allows a single observed context \(\mathbf{c}_t\) to induce different feature
vectors for different arms. In our implementation, \(\phi(\cdot)\) concatenates
task-relevant context cues with any arm-specific indicators needed to model
arm-dependent reward parameters. The exploration coefficient \(\alpha\) in Algorithm~\ref{alg:linucb} tunes the optimism in the upper
confidence bound. In practice, \(\alpha\) is chosen empirically (or via a
theoretical schedule) to balance exploitation of high-value arms and
exploration of uncertain arms. 
\begin{algorithm}[t]
\caption{Disjoint LinUCB for Contextual Multi-Armed Bandits}
\label{alg:linucb}
\KwIn{
Arms $\mathcal{A}=\{1,\dots,A\}$,
regularization $\lambda>0$,
exploration coefficient $\alpha>0$,
feature map $\phi(\mathbf{c}_t,a)\in\mathbb{R}^d$
}
\KwInit{
For each arm $a\in\mathcal{A}$:\;
\hspace{1em}$\mathbf{A}_a \leftarrow \lambda \mathbf{I}_d$,\quad
$\mathbf{b}_a \leftarrow \mathbf{0}_d$
}
\For{$t=1,2,\dots,T$}{
Observe context $\mathbf{c}_t$ and admissible arms $\mathcal{A}_t\subseteq \mathcal{A}$\;
\ForEach{$a\in \mathcal{A}_t$}{
$\mathbf{x}_{t,a} \leftarrow \phi(\mathbf{c}_t,a)$\;
$\hat{\boldsymbol{\theta}}_a \leftarrow \mathbf{A}_a^{-1} \mathbf{b}_a$\;
$p_{t,a} \leftarrow \mathbf{x}_{t,a}^\top \hat{\boldsymbol{\theta}}_a
\;+\; \alpha \sqrt{\mathbf{x}_{t,a}^\top \mathbf{A}_a^{-1} \mathbf{x}_{t,a}}$\;
}
Select arm:
$a_t \leftarrow \arg\max_{a\in \mathcal{A}_t} p_{t,a}$\;
Execute arm $a_t$ and observe reward $r_t$\;
Update chosen arm:\;
\hspace{1em}$\mathbf{A}_{a_t} \leftarrow \mathbf{A}_{a_t} + \mathbf{x}_{t,a_t}\mathbf{x}_{t,a_t}^\top$\;
\hspace{1em}$\mathbf{b}_{a_t} \leftarrow \mathbf{b}_{a_t} + r_t \mathbf{x}_{t,a_t}$\;
}
\end{algorithm}

The reward used by the CMAB policy is defined as a weighted combination of the contextual state variables,
\begin{equation}
r_t
=
w_1 \dot{m}_t^{\rm occ}
-
w_2 B_t
+
w_3 V_t^{\rm vlm}
+
w_4 V_t^{\rm sim}
+
w_{succ} \ ,
\label{eq:reward}
\end{equation}
where \(w_i\) denotes the weight associated with each state feature, and $w_{succ}$ acts as a terminal task success reward. This reward formulation encourages actions that increase map coverage, improve semantic relevance, and favor confident VLM predictions, while discouraging unnecessary revisitation of previously explored regions. The backtracking indicator \(B_t\) is the only negatively weighted term, penalizing goal positions that require retracing past exploration paths. The weights for this reward function are given in Table \ref{tab:reward_weights}.

\begin{table}[t]
\centering
\caption{Reward components and corresponding weights used by the CMAB policy.}
\label{tab:reward_weights}
\begin{tabular}{lc}
\toprule
Reward Component & Weight ($w$) \\
\midrule
Map coverage rate ($\dot{m}_t^{\rm occ}$) & 1.0 \\
Backtracking ($B_t$) & -1.0 \\
VLM confidence ($V_t^{\rm vlm}$) & 1.0 \\
Similarity weight & 1.0 \\
Task success & 10.0 \\
\bottomrule
\end{tabular}
\end{table}

\subsection{OpenRoboVox}
\label{methods:openRoboVox-implementation}

\begin{figure*}[!t]
    \centering
    \includegraphics[width=\textwidth]{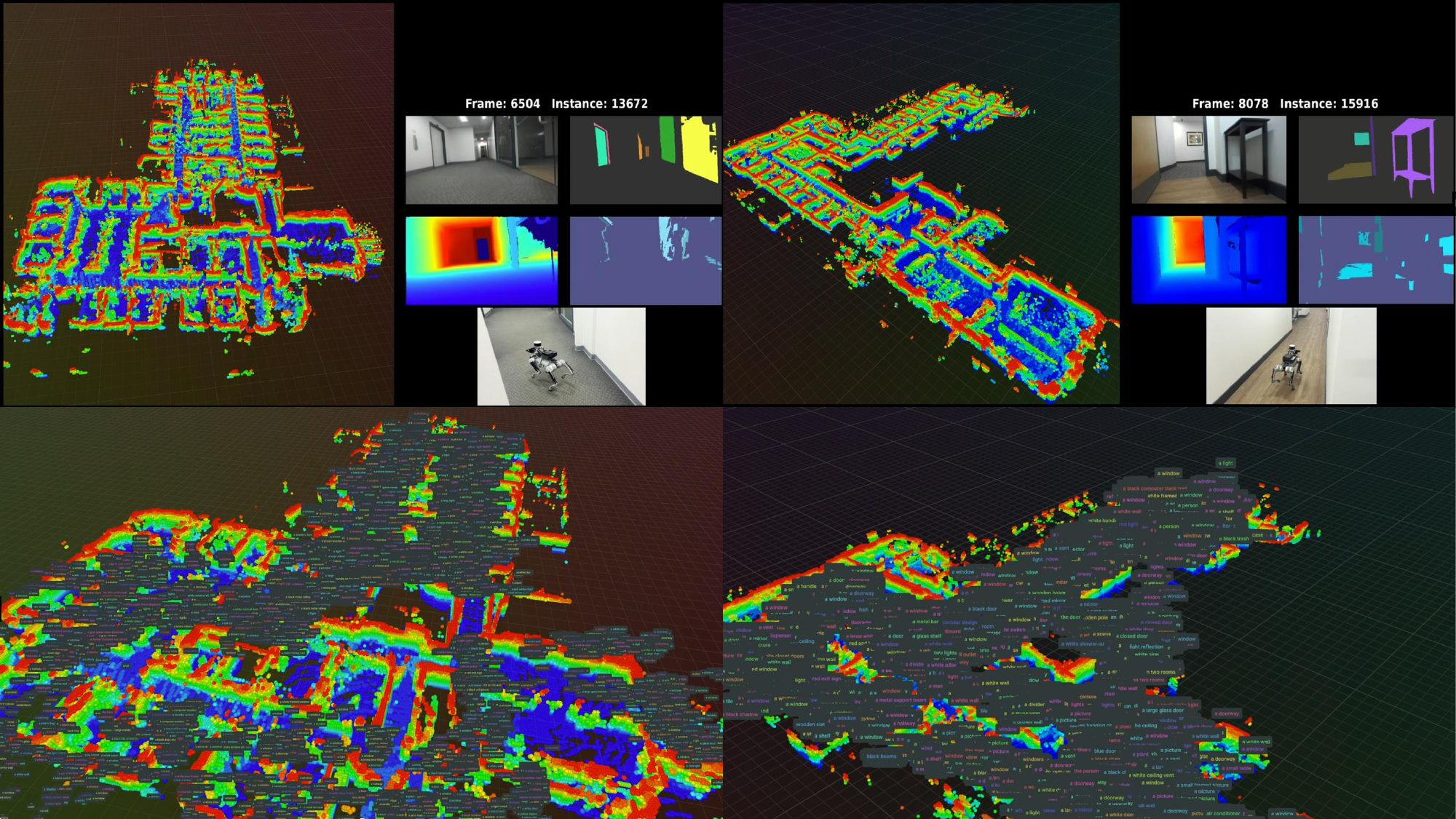}
    \captionsetup{font=small}
    \caption{\textbf{\textit{OpenRoboVox} Hardware Validation.} Top row shows the 3D occupancy grid and the OpenRoboVox framework, including the RGB and Depth camera streams, semantic segmentation, and corresponding semantic voxels. The bottom row shows the 3D occupancy, overlaid by the captions of the scene dictionary, for two different floors of an office building (left column is for floor 1 and right column for floor 2).}
    \label{fig:octomap_hardware_demo}
\end{figure*}

Our \textit{OpenRoboVox} framework builds upon the architecture of OpenVox~\cite{deng2025openvoxrealtimeinstancelevelopenvocabulary}. While OpenVox introduces an effective method for instance-level mapping using probabilistic voxels, it is primarily validated in the Habitat simulator, on offline datasets, or on small-scale maps in simplified hardware experiments. Such settings provide near-perfect state estimation, depth estimation and assume abundant computational resources, masking the critical bottlenecks faced during deployment on physical hardware. Here, we transition the framework from simulation or ideal environments to reality, deploying it on a Unitree Go2 mobile robot. This shift necessitates a fundamental re-engineering of the core pipeline to handle the constraints of GPU, noisy real-world odometry, and the requirement for continuous, large-scale operation.

The OpenVox framework~\cite{deng2025openvoxrealtimeinstancelevelopenvocabulary} integrates a geometric back-end with a semantic front-end. It uses Truncated Signed Distance Fields (TSDF) for geometry and a probabilistic voxel grid for instance semantics, where each voxel $v$ maintains a Dirichlet posterior $\boldsymbol{\theta}_v$ over the set of global instance IDs $\Gamma_t$. The perception pipeline leverages YOLO-World \cite{cheng2024yoloworld} for detection and the Tokenize Anything Model (TAP) \cite{pan2024tokenize} for segmentation, fusing these outputs into 3D space via Bayesian updates. However, deploying this architecture directly on a mobile robot revealed three critical challenges:

\begin{itemize}
    \item  The original map expansion strategy duplicates the entire global tensor during resizing events, doubling peak VRAM usage. In our simulation and hardware deployments, this caused out-of-memory (OOM) crashes within minutes of several GPUs, including the most powerful GPU evaluated, the RTX 4090.
    \item Concurrency Locking: The sequential execution of the perception stack (YOLO-World + TAP) blocks the geometric tracking thread. On a mobile robot, the resulting latency (\SI{>200}{\ms} per cycle) breaks the real-time odometry chain, causing significant map drift and delayed map updates.
    \item Lack of Queryable Semantics: While the voxels are mapped to instance IDs, there is no mechanism or data structure for efficient high-level semantic queries or spatial relationships (e.g., ``Where is the red chair?") without scanning millions of voxels per frame.
\end{itemize}

To resolve these issues, \textit{OpenRoboVox} introduces three architectural changes: (i) a Scene-Dictionary manager, an incremental background process that decouples semantic reasoning from geometric tracking; (ii) a block-based volumetric transfer protocol that bounds peak VRAM during map expansion; and (iii) an asynchronous parallel architecture, a decoupled ROS2 design that preserves geometric stability regardless of semantic latency. The result of \textit{OpenRoboVox} is visually demonstrated in Fig. \ref{fig:octomap_hardware_demo}.

\subsubsection{Scene-Dictionary Manager}
\label{methods:scene_dictionary}
The Scene-Dictionary $\mathcal{S}_t$ bridges the gap between raw voxel data and high-level planning, and operates asynchronously from the main SLAM thread. In the baseline implementation, extracting semantic objects requires a global aggregation function $f_{\textrm{agg}}$ over the entire set of map instances $\Gamma_{t}$, such that the computational cost is $C_{\textrm{total}} \propto |\Gamma_{t}|$. As the map grows $|\Gamma_{t}|$ quickly exceeds the per-frame instance budget $N_{\textrm{budget}}$, and aggregation can no longer meet the real-time constraint.

We resolve this via an incremental background manager that employs a priority-based scheduling algorithm. Let $\Omega_t \subset \Gamma_t$ be the subset of instances selected for processing at cycle $t$. The Scene-Dictionary update is the partial mapping
\begin{equation}
\mathcal{S}_{t+1} = \mathcal{S}_t \cup \{ \text{Extract}(O_i) \mid O_i \in \Omega_t \} \ ,
\end{equation}
where $\text{Extract}(\cdot)$ represents the geometric parameter estimation (centroid, bounding box, occupancy probability) from the voxel map. To determine $\Omega_t$, we employ a Priority-Based Scheduling Algorithm. Instead of refreshing the entire map, the manager selects a bounded batch $\Omega_t$ (where $|\Omega_t| \ll |\Gamma_t|$) based on a priority union
\begin{equation}
\Omega_t = \mathcal{P}_{rec} \cup \mathcal{P}_{vis} \cup \mathcal{P}_{rot} \ ,
\end{equation}
including:
\begin{itemize}
\item Temporal Recency ($\mathcal{P}_{rec}$): Instances updated by the tracking thread where $(t_{now} - t_{last\_seen}) < \tau_{rec}$.
\item Active Observation ($\mathcal{P}_{vis}$): Instances where the projected voxel set $V_{O_i}$ intersects with the current camera frustum $\mathcal{F}_t$.
\item Rotational Consistency ($\mathcal{P}_{rot}$): A subset of static, older instances $\{O_k\}$ selected via a round-robin index $k \equiv t \pmod K$ to ensure global graph consistency.
\end{itemize}
Because $|\Omega_t|$ is bounded by design, the per-cycle update cost is $O(1)$ with respect to the map size $|\Gamma_t|$. 
We further accelerate the $\text{Extract}(\cdot)$ operator through batched mask computation: voxel masks for all instances in $\Omega_t$ are stacked into a single tensor $\mathbf{W}_t$, and instance properties are computed by a vectorized operator $\Phi_{\textrm{mask}}(\mathbf{W}_t)$ rather than per-instance loops. This avoids redundant allocation of per-instance coordinate grids and reduces semantic update latency by roughly $90\%$ in our measurements.

\subsubsection{GPU Optimization \& Memory Safety}
\label{methods:gpu_optimization}
Continuous large-scale mapping requires bounded VRAM usage, since the map grows unboundedly while GPU memory is fixed. The dominant bottleneck is map expansion: standard volumetric mapping resizes the global domain $\mathcal{D}$ to $\mathcal{D}'$ through contiguous reallocation, briefly duplicating the entire map and giving a peak demand $M_{\textrm{peak}} \approx |\mathcal{V}_{\mathcal{D}}| + |\mathcal{V}_{\mathcal{D}'}|$. We reduce the peak demand with a partitioned transfer protocol that decomposes $\mathcal{D}'$ into disjoint cubic blocks $\mathcal{B} = \{b_1, \dots, b_{N_b}\}$ of size $B^3$ ($B=64$ in our implementation) and transfers them one at a time
\begin{equation}
\mathcal{V}_{\mathcal{D}'} = \bigcup_{k=1}^{N_b} \mathcal{T}(\mathcal{V}_{\mathcal{D}} \cap b_k).
\end{equation}
This bounds the peak overhead to the global map size plus a single block buffer, $M_{\textrm{peak}} \approx |\mathcal{V}_{\mathcal{D}'}| + |b_{\textrm{buffer}}|$ with $|b_{\textrm{buffer}}| \propto B^3$, flattening the allocation curve and preventing the out-of-memory failures observed during boundary expansion. We additionally apply standard inference-time memory hygiene, i.e., detaching tensors from the autograd graph and releasing allocations each cycle, to keep the footprint stable across long runs.

\subsubsection{Asynchronous Parallel Architecture}
\label{methods:parallel_architecture}
Geometric integration runs at $\sim$15\,Hz, while the open-vocabulary perception stack (YOLO-World + TAP) runs at $\sim$5\,Hz. To reconcile these timescales, \textit{OpenRoboVox} runs the two pipelines on independent threads.
This architecture incorporates a custom concurrency control manager that enforces an isolation boundary between the high-bandwidth tracking processes and the latency-prone semantic processes. We utilize fine-grained mutual exclusion primitives to govern access to the shared global instance registry, effectively decoupling the two streams:
\begin{itemize}
    \item Geometric Stream: The tracking thread performs voxel occupancy updates with negligible locking overhead (sub-millisecond critical sections). This ensures that the odometry chain remains non-blocking and strictly adheres to real-time constraints.
    \item Semantic Stream: The perception thread operates asynchronously, effectively masking inference latency. Resource locks are acquired solely during the terminal data fusion stage, allowing heavy computation to proceed without interrupting the geometric update cycle.
\end{itemize}

\subsection{2D Pillar Map \& Similarity Maps}\label{methods:2d_pillar}

While \textit{OpenRoboVox} provides large-scale 3D probabilistic representation, utilizing raw voxel data for high-level navigation planning is computationally prohibitive. A mobile robot operating primarily in the $SE(2)$ space does not require full volumetric semantic density for path planning. Instead it requires a projected understanding of navigability and semantic relevance. Performing high-dimensional feature comparisons (e.g., cosine similarity) across thousands of voxels for every navigation step introduces unnecessary latency. To address this, we introduce the 2D Pillar Map, a height-compressed, instance-labeled grid structure that enables efficient computation of semantic potential fields, specifically Target Object Similarity ($\mathcal{T}_{\textrm{sim}}$), Neighbor Similarity ($\mathcal{N}_{\textrm{sim}}$), and Constraint Similarity ($\mathcal{C}_{\textrm{sim}}$).

\subsubsection{2D Pillar Map Construction}
\label{sec:2d-pillar}
Let $\mathcal{V} = \{v_{xyz}\}$ denote the voxel set produced by \textit{OpenRoboVox}, where each voxel $v$ stores its Dirichlet posterior $\boldsymbol{\theta}_v$ over instance IDs. We define the 2D Pillar Map as a projection $\Pi: \mathbb{R}^3 \to \mathbb{R}^2$ that collapses the 3D voxel space $\mathcal{V}$ onto a 2D grid $\mathcal{G}_t$ at resolution $r$. The \textit{pillar} $\Pi_{uv}$ at cell $(u, v)$ is the set of occupied voxels that project to $(u, v)$
\begin{equation}
\Pi_{uv} = \{ v_{xyz} \in \mathcal{V} \mid \lfloor x/r \rfloor = u, \lfloor y/r \rfloor = v, \rho(v_{xyz}) > \tau_{\textrm{occ}} \},
\end{equation}
where $r$ is the grid resolution, and $\rho(\cdot)$ represents the occupancy probability of voxel $v_{xyz}$, and $\tau_{\textrm{occ}}$ is the occupancy threshold.

Unlike semantic occupancy grids that typically store a single instance class or instance score within a single grid cell, our Pillar Map maintains a ranked hierarchy of the top-$K$ dominant instances. This stratification preserves critical vertical context, such as differentiating a target object resting on top of a support surface, thereby retaining instance-level granularity within a computationally efficient 2D projection.

Formally, let $\Gamma_t$ denote the set of global instance IDs and let $I(v') = \arg\max_\gamma \boldsymbol{\theta}_{v'}[\gamma]$ denote the Maximum A Posteriori (MAP) instance assignment for voxel $v'$ under its Dirichlet posterior $\boldsymbol{\theta}_{v'}$ (Sec.~\ref{methods:openRoboVox-implementation}). For each instance $\gamma \in \Gamma_t$, its voxel count within pillar $\Pi_{uv}$ is
\begin{equation}
    \omega_\gamma(u,v) = \sum_{v' \in \Pi_{uv}} \mathbbm{1}\!\left(I(v') = \gamma\right).
\end{equation}
The cell state $S_{uv}$ is the top-$K$ instances ordered by voxel count
\begin{equation}
    S_{uv} = \big((\gamma_1, \omega_1), \ldots, (\gamma_K, \omega_K)\big), \quad \omega_1 \geq \omega_2 \geq \cdots \geq \omega_K,
\end{equation}
with $K=4$ in our implementation. The projection is recomputed in a background thread so that pillar updates remain current with the latest voxel state without blocking geometric integration.

\subsubsection{Target Object(s) Similarity ($\mathcal{T}_{\textrm{sim}}$)}
\label{methods:similarity}
To enable semantic navigation (e.g., ``Go to the red chair"), we compute a dense similarity potential field, denoted as $\mathcal{T}_{\textrm{sim}}$. This process relies on the precomputed text-embeddings of the system without incurring the cost of querying an LLM or VLM for every frame. Let $\mathcal{Q}_{\mathcal{T}}$ be the natural language query describing the target. We encode this query into a high-dimensional embedding vector $\mathbf{f}_{\mathcal{Q}_{\mathcal{T}}} \in \mathbb{R}^{384}$ using the shared SBERT encoder. Similarly, every instance $\gamma \in \Gamma_t$ in the global map possesses a semantic feature vector $\mathbf{f}_\gamma$, which is the aggregated result of the fusion of caption embeddings over time. The Target Object(s) Similarity map $\mathcal{T}_{\textrm{sim}} \in \mathbb{R}^{H \times W}$ is computed by projecting the semantic relevance of 3D instances onto the 2D Pillar grid as mentioned in Section~\ref{sec:2d-pillar}. For every cell $(u, v)$ in the grid, the similarity score is defined as the maximum cosine similarity among all instances contained within that pillar
\begin{equation}
\label{eq:sim_score}
\mathcal{T}_{\textrm{sim}}(u, v) = \max_{(\gamma, \omega) \in S_{uv}} \left( \frac{ \mathbf{f}_{\mathcal{Q}_{\mathcal{T}}} \cdot \mathbf{f}_\gamma}{\| \mathbf{f}_{\mathcal{Q}_{\mathcal{T}}} \| \| \mathbf{f}_\gamma \|} \right).
\end{equation}
By utilizing the max-pooling operator over the pillar's instance set $S_{uv}$, the system ensures that a small target object (high similarity) located on top of a large background object (low similarity) is effectively prioritized as the dominant semantic signal in the $\mathcal{T}_{\textrm{sim}}$ map.

\subsubsection{Other Similarity Maps}
Beyond the primary target object $\mathcal{Q}_{\mathcal{T}}$, semantic navigation often relies on spatial context (e.g., ``near the table'') or negative constraints (e.g., ``avoid the wet floor''). To support these requirements, we extend the similarity computation to generate two additional potential fields:

\textbf{Neighbor Similarity ($\mathcal{N}_{\text{\textrm{sim}}}$)}:
The definition of a target often includes spatial relationships with other objects. Let $\mathcal{Q}_{\mathcal{N}} = \{n_1, n_2, \dots, n_m\}$ be a set of expected neighbor objects in the query (e.g., $\{$``table'', ``desk''$\}$ when looking for a chair). We compute the Neighbor Similarity map $\mathcal{N}_{\text{\textrm{sim}}}$ by aggregating the semantic relevance of all neighbor prompts. For a pillar at $(u, v)$, the score is the maximum similarity between any instance in the pillar and any prompt in the neighbor set
\begin{equation}
    \mathcal{N}_{\text{\textrm{sim}}}(u, v) = \max_{(\gamma, \omega) \in S_{uv}} \left( \max_{q \in \mathcal{Q}_{\mathcal{N}}} \frac{\mathbf{f}_{q} \cdot \mathbf{f}_\gamma}{\| \mathbf{f}_{q} \| \| \mathbf{f}_\gamma \|} \right).
    \label{eq:neighbor_similarity}
\end{equation}
This map identifies regions rich in contextual evidence, which is subsequently used to filter the Scene-Dictionary for the LLM (see Sec.~\ref{methods:semantic_reasoning}).

\textbf{Constraint Similarity ($\mathcal{C}_{\text{\textrm{sim}}}$):}
We also define a set of negative constraints $\mathcal{Q}_{\mathcal{C}}  = \{c_1, c_2, \dots\}$. The Constraint Similarity map $\mathcal{C}_{\text{\textrm{sim}}}$ highlights regions containing objects that must be strictly avoided during path planning. Calculated identically to Eq.~\eqref{eq:neighbor_similarity}, this map operates as a dynamic ``semantic cost layer'' in the local costmap, inflating the obstacle radius around specific semantic classes (e.g., fragile objects, hazardous areas, or people) without altering the geometric occupancy grid.

\subsection{Frontier Candidate Generation and Goal Selection}
\label{methods:frontier_generation}

Let $F_t$ be the free (known-clear) cells and $U_t$ be the unknown cells in the occupancy grid. Frontier cells are:
\begin{equation}
\mathcal{F}_t^{\mathrm{frontier}}
:=
\left\{
v \in F_t
\;\middle|\;
\exists\, u \in U_t,\;
\mathcal{N}(u)\cap \mathcal{N}(v)\neq\emptyset
\right\}.
\end{equation}
where $\mathcal{N}(x)$ denotes the neighborhood of cell $x$, and we cluster $\mathcal{F}_t^{\text{frontier}}$ into a finite set of representative frontier centers $\mathcal{C}_t \subseteq F_t$.

For a candidate goal $c \in \mathcal{C}_t$, define the information gain
\begin{equation}
I_t(c) := \lvert U_t \cap \mathcal{S}(c) \rvert,
\end{equation}
where $\mathcal{S}(c)$ is the sensor footprint (cells observable from $c$). Among candidates admitting an energy-feasible plan (i.e., a shortest path from the robot to $c$ and then to a known charging station within the available energy budget), we choose
\begin{equation}
\mathbf{g}_t \in \arg\min{c \in \mathcal{C}_t}\Big( -I_t(c) + \alpha_f,\ell_t(c)\Big),
\end{equation}
where $\ell_t(c)$ is the corresponding path length and infeasible candidates are assigned infinite cost. Further details of the frontier selection algorithm are provided in \cite{Karumanchi2025energyConstrained}

\subsection{Semantic Map Goal Selection}
\label{methods:semantic_reasoning}

We implement an asynchronous query interface that bridges the robot's perception system with an LLM backend running in a background worker thread. Our main experiments use GPT-4o; we also validate the framework with the smaller Qwen2.5-VL-7B model. To maintain real-time performance, the worker is invoked only when the Scene-Dictionary changes.

Whenever the priority batch $\Omega_t$ updates one or more instances, the worker regenerates a textual snapshot of the world state. Let the context $\mathbf{C}_t$ be the serialized representation of the Scene-Dictionary at time $t$. The serialization function $\Psi: \mathcal{S} \to \Sigma^*$ applies truncation to fit within the token budget $N_{\max}$ (e.g. 128k tokens for GPT-4o)
\begin{equation}
\mathbf{C}_t = \Psi(\mathcal{S}_t) = \text{Header}(\mathcal{S}_t) \oplus \bigoplus_{O_i \in \text{Sort}(\mathcal{S}_t)} \text{Desc}(O_i).
\label{eq:context_serialization}
\end{equation}

Here, $\text{Sort}(\cdot)$ ranks instances in descending order of their occupancy probability $\kappa_i$ to ensure the most reliable landmarks are always preserved in the prompt, while $\text{Desc}(O_i)$ formats individual instance metadata (ID, caption, centroid $\mathbf{p}_i$, and bounding box $\mathbf{b}_i$).

A brute-force injection of the entire Scene-Dictionary $\mathcal{S}_t$ into the LLM context is inefficient and would exceed the token budget ($N_{\max}$) for large-scale environments. To resolve this, we employ a pre-filtering stage based on the semantic potential fields defined in Sec.~\ref{methods:similarity}. We define a filtered subset $\tilde{\mathcal{S}}_t \subset \mathcal{S}_t$ containing only instances relevant to the current query $Q$. An instance $O_i$ is retained if it satisfies either the target or neighbor similarity thresholds
\begin{equation}
\begin{aligned}
O_i \in \tilde{\mathcal{S}}_t \iff{} & \left( \frac{\mathbf{f}_{O_i} \cdot \mathbf{f}_{\mathcal{Q}_{\mathcal{T}}}}{\|\mathbf{f}_{O_i}\| \, \|\mathbf{f}_{\mathcal{Q}_{\mathcal{T}}}\|} > \delta_{\mathcal{T}} \right) \\
& \lor \left( \exists q \in \mathcal{Q}_{\mathcal{N}},\, \frac{\mathbf{f}_{O_i} \cdot \mathbf{f}_q}{\|\mathbf{f}_{O_i}\| \, \|\mathbf{f}_q\|} > \delta_{\mathcal{N}} \right),
\end{aligned}
\label{eq:context_filtering}
\end{equation}
where $\delta_{\mathcal{T}}$ and $\delta_{\mathcal{N}}$ are configurable similarity thresholds. Adjusting these parameters modulates the trade-off between embedding-based retrieval and LLM-based reasoning: higher thresholds are faster but rely strictly on the discriminative power of the feature space, while lower thresholds expand the search context and let the LLM resolve ambiguities.

For navigation scoring, we construct a filtered prompt context $\tilde{\mathbf{C}}_t = \Psi(\tilde{\mathcal{S}}_t)$ from the filtered subset $\tilde{\mathcal{S}}_t$, reusing the same serialization $\Psi$ as in Eq.~\ref{eq:context_serialization}. We further augment the textual description of each instance with its computed similarity scores, $\boldsymbol{\sigma}_i = [\sigma_{i}^{\text{targ}}, \sigma_{i}^{\text{neigh}}]^\top$, representing the instance's alignment with the semantic search criteria.

By explicitly embedding the cosine similarity metrics into the prompt, we ground the LLM's reasoning in the underlying vector space, allowing it to weigh the visual semantic evidence (via $\boldsymbol{\sigma}_i$) against the spatial geometric evidence (via coordinates and neighbors).

The LLM is tasked with ranking candidate instances according to both semantic relevance and spatial plausibility. Specifically, we define the scene-aware query $Q_{\text{scene}}$ as a request for the LLM to rank filtered instances based on (i) semantic match and (ii) spatial plausibility. The LLM returns a structured JSON response containing a refined confidence score $\Phi_{\text{LLM}}(O_i) \in [0,1]$ for each candidate
\begin{equation}
    \Phi_{\text{LLM}}(O_i) = \text{LLM}\!\big(O_i, \mathcal{N}_r(O_i) \mid \mathcal{Q}_{\mathcal{T}},\, \mathcal{Q}_{\mathcal{N}},\, \boldsymbol{\sigma}_i\big),
    \label{eq:llm_scoring}
\end{equation}
where $\mathcal{N}_r(O_i)$ represents the set of instances within a radius $r_{\mathcal{N}}$ of $O_i$.

The scoring mechanism effectively suppresses high-confidence false positives by enforcing local semantic consistency. Thus, the semantic map expert used in the bandits algorithm retains the top $N$ instances according to $\Phi_{\text{LLM}}$, and selects the final navigation goal as
\begin{equation}
    \mathbf{g}_t = \mathbf{p}_{i^\ast}, \quad i^\ast = \arg\max_{O_i \in \text{Top-}N} \Phi_{\text{LLM}}(O_i),
\end{equation}
where $\mathbf{p}_{i^\ast}$ corresponds to the centroid location of the highest-scoring instance among the top-$N$ candidates.

The full prompt to the LLM combines the system prompt $\mathcal{I}_{\text{sys}}$ defining the agent's role (Fig.~\ref{fig:llm_prompt}), the filtered scene context $\tilde{\mathbf{C}}_t$, the user querey, $\mathcal{L}_t$, and optionally the egocentric scene descriptions $D_{\text{scene}}$ accumulated by the VLM expert during its goal selection procedure (see Sec.~\ref{methods:vlm_goal_selection}).

\subsection{Egocentric VLM Goal Selection}
\label{methods:vlm_goal_selection}

\begin{figure*}[!t]
    \centering
    
    \begin{subfigure}{\textwidth}
        \centering
        \includegraphics[width=\textwidth]{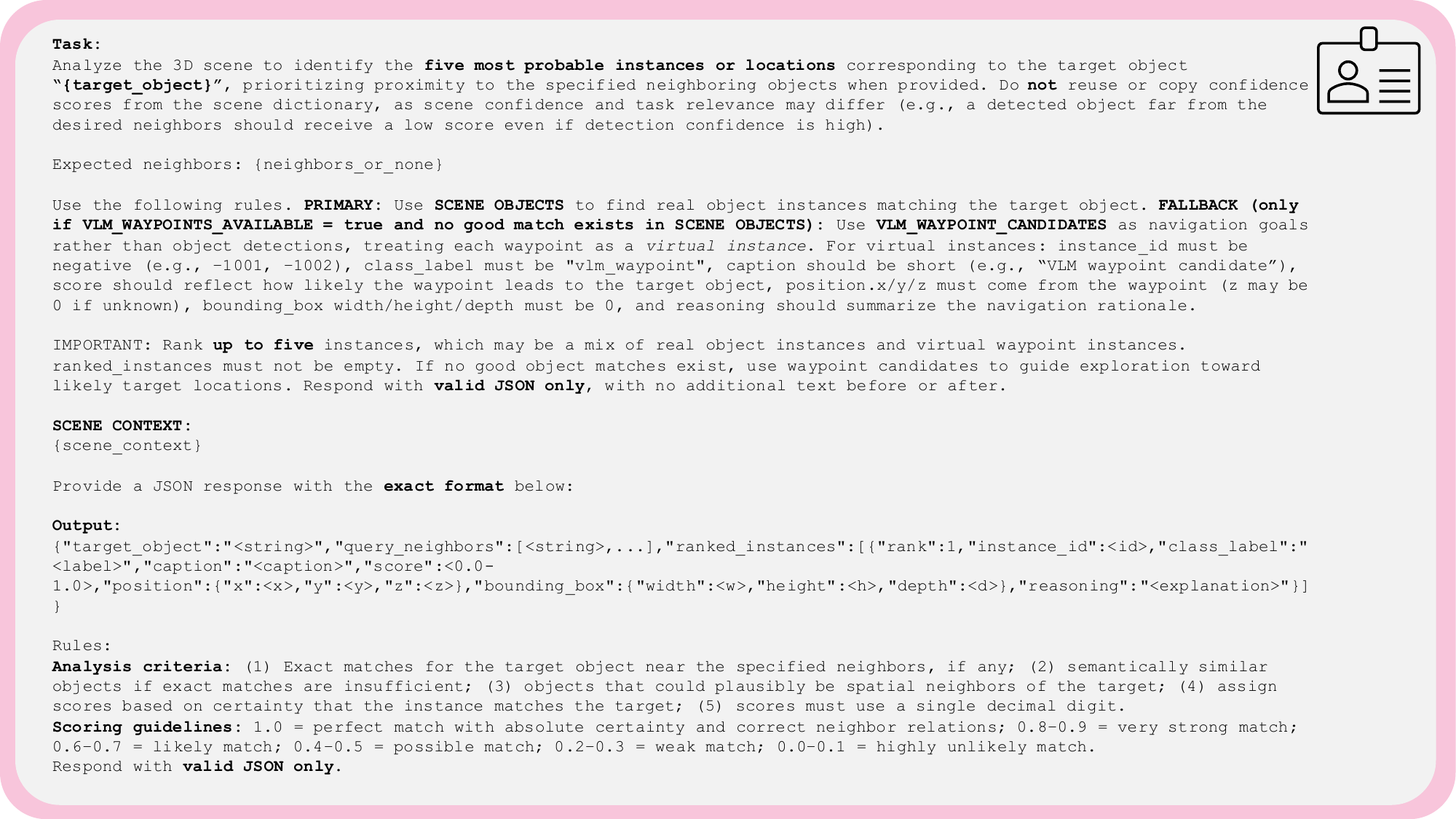}
        \caption{Input system prompt for the semantic map expert.}
        \label{fig:llm_prompt}
    \end{subfigure}
    
    \vspace{1em}
    
    \begin{subfigure}{\textwidth}
        \centering
        \includegraphics[width=\textwidth]{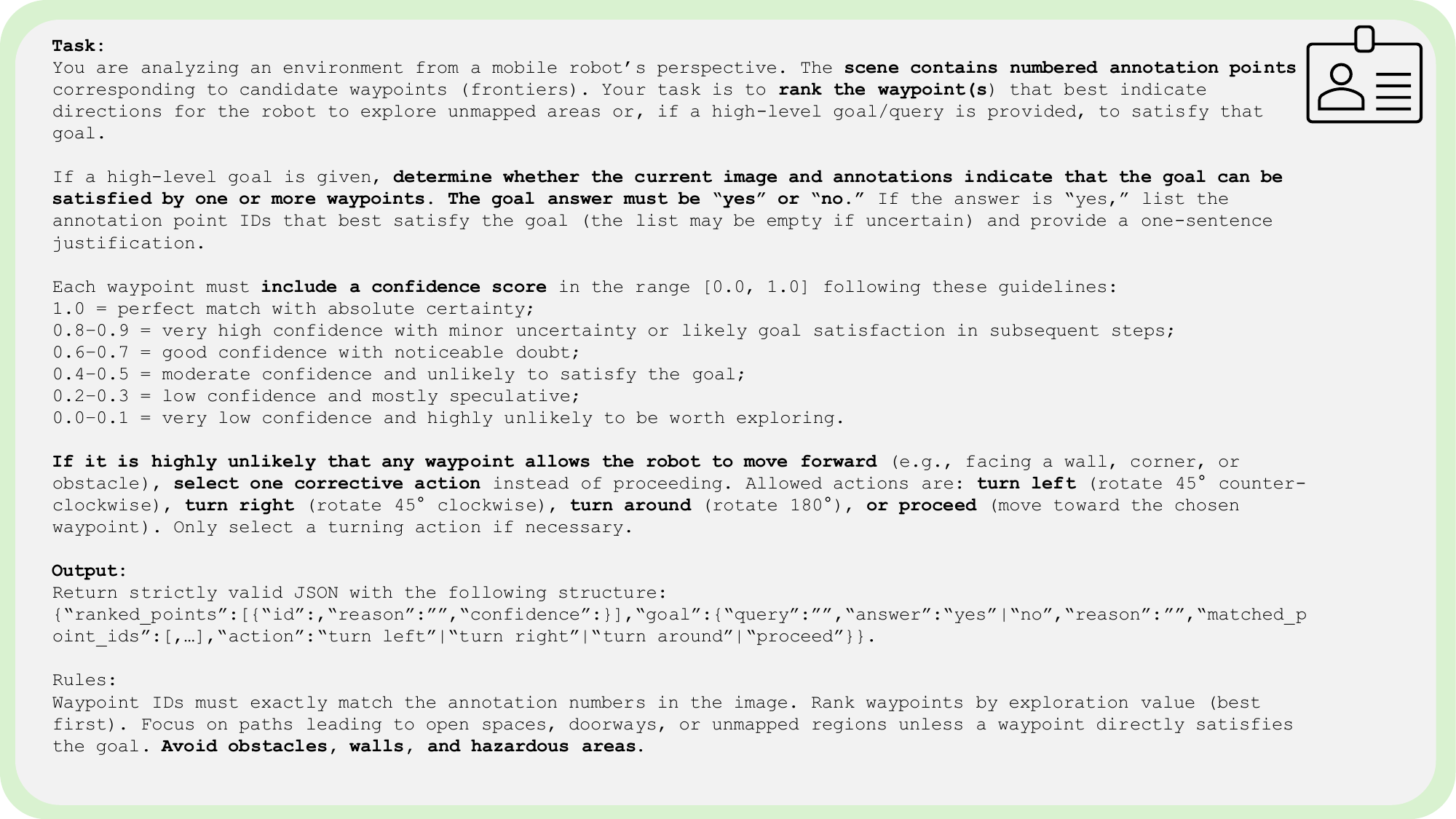}
        \caption{Input system prompt for the egocentric VLM expert.}
        \label{fig:vlm_prompt}
    \end{subfigure}
    
    \captionsetup{font=small}
    \caption{\textbf{Input system prompts.} System prompts used for the semantic map and ego-centric VLM experts for validation experiments.}
    \label{fig:input_prompts}
\end{figure*}

The VLM inputs consist of a user query, $\mathcal{L}_t$, a system prompt $\mathcal{I}_{\mathrm{sys}}$ (Fig.~\ref{fig:vlm_prompt}), and an RGB image $I_t$ of the scene. The image is augmented with spatial annotations derived from an evenly spaced grid over the image plane. Annotations corresponding to grid cells that are either too close to the robot or beyond a maximum interaction range are pruned prior to inference.

Similar to Sec.~\ref{methods:semantic_reasoning}, we employ an asynchronous VLM worker thread that submits queries to a GPT-4o (or Qwen2.5-VL-7B) model backend and continuously monitors for completed responses, enabling non-blocking integration with the planning pipeline.

The VLM returns a structured JSON response that includes (i) confidence scores $\phi_{\mathrm{VLM}}(O_i) \in [0,1]$ for each annotated candidate $O_i$, and (ii) a high-level action selection
\(
a_t \in \mathcal{A},
\)
where the discrete action set is defined as
\begin{equation}
\mathcal{A} := \{\actProceed,\ \actLeft,\ \actRight,\ \actTurn\}.
\end{equation}
Here, $\actProceed$ denotes proceeding toward an annotated goal, $\actLeft$ and $\actRight$ denote in-place left and right rotations, and $\actTurn$ denotes a $180^\circ$ turn.

When the selected action is $\actProceed$, the navigation goal is chosen according to
\begin{equation}
\mathbf{g}_t = \arg\max_{O_i} \; \phi_{\mathrm{VLM}}(O_i),
\end{equation}
where $\mathbf{g}_t$ corresponds to the spatial location of the highest-confidence instance. If instead the VLM selects a turning action, the annotation-based goal selection is overridden and the robot executes the corresponding discrete motion primitive.

To ensure stable behavior, VLM-based decisions are executed over a finite planning horizon. Specifically, when a VLM action is selected, we roll out a sequence of VLM decisions over a horizon of length $H$, where $H=3$ in our implementation. Each element of the sequence may correspond either to a navigation goal (via $\actProceed$) or to a discrete turning action, and each turning action consumes one step of the horizon. The final navigation goal $\mathbf{g}_t$ is thus obtained either directly from the current VLM output or after executing up to $H$ chained VLM actions.

Finally, the VLM-generated reasoning used to interpret the current scene is stored in a scene-description dictionary, $D_{\rm scene}$. Because the VLM reasons over annotations that include the associated 3D coordinates, the scene descriptions capture both semantic and spatial context. The information is subsequently provided to the semantic map expert (Sec. \ref{methods:semantic_reasoning}) to enrich contextual understanding and support more informed decision-making.



\section{Experimental Validation}
\label{sec:experimental_validation}

\subsection{Simulation and Hardware Setup}
\label{simulation_hardware_setup}
We evaluate \textit{RoboAtlas} on three platforms: Isaac Sim~\cite{NVIDIA_Isaac_Sim} environment, the photo-realistic Habitat simulator~\cite{savva2019habitat, ramakrishnan2021hm3d}, and hardware with a Unitree Go2 quadruped. In all three settings, the \textit{RoboAtlas} decision and mapping stack described in Sec.~\ref{methods:main_sec} runs on a desktop with an NVIDIA RTX 4090 GPU. Our main experiments use GPT-4o, accessed via Microsoft Azure, for all VLM and LLM tasks. We additionally evaluate the smaller Qwen2.5-VL-7B as a foundation-model variant. The platforms differ in how the robot-side perception, localization, and actuation are realized.

\begin{figure*}[t]
    \centering
    \includegraphics[width=0.99\textwidth]{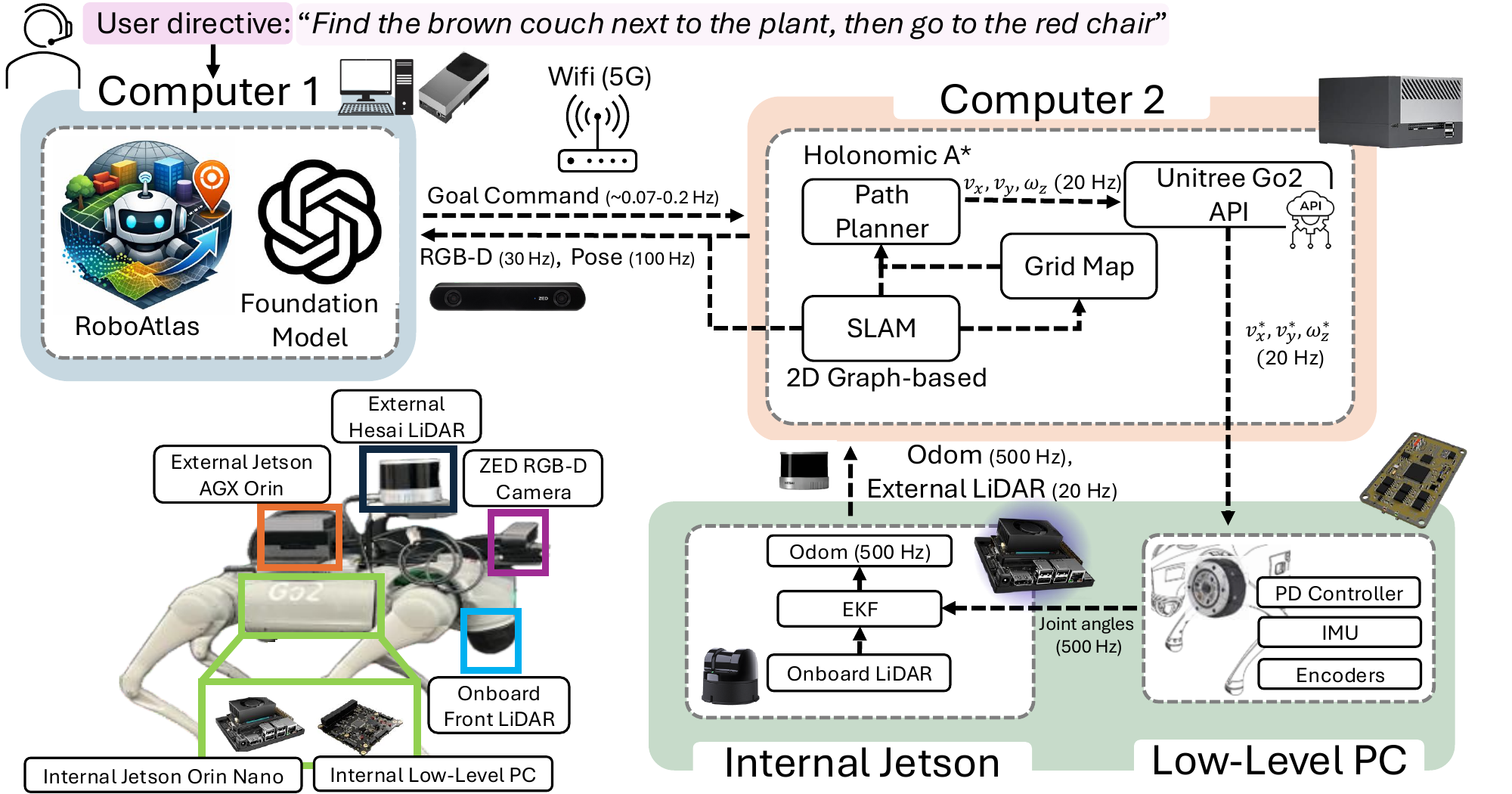}
    \captionsetup{font=small}
    \caption{\textbf{Overall Hardware System Flowchart.} The user provides a high-level language directive, which is processed by \textit{RoboAtlas} and a foundation model running on an external desktop GPU. RGB-D observations and robot pose estimates are streamed from the Unitree Go2 platform through an internal Jetson AGX Orin to the desktop, where semantic mapping, contextual reasoning, and goal selection are performed. \textit{RoboAtlas} generates navigation goals that are transmitted back to the robot via WiFi. Onboard, the ROS2-based navigation stack integrates SLAM Toolbox for localization and mapping, Nav2 with a holonomic A* planner for path planning, and the Unitree API for motion execution. Low-level control combines odometry, IMU, encoder, and LiDAR measurements through an EKF to provide robust state estimation and closed-loop navigation in real-world environments.}
    \label{fig:system_flowchart}
\end{figure*}

\textbf{Hardware.} As shown in Fig.~\ref{fig:system_flowchart}, the desktop communicates over WiFi (5\,GHz) with an onboard Jetson AGX Orin mounted on the robot. The Jetson runs the Active SLAM stack under ROS2, using SLAM Toolbox~\cite{Macenski2021} for localization and 2D grid mapping and Nav2~\cite{macenski2023survey} for autonomous navigation. We use the Nav2 SMAC planner, which performs a holonomic A* search over the occupancy grid while modeling the robot as a rectangular footprint for collision avoidance~\cite{macenski2024smac}. Odometry at 500\,Hz is obtained from the Unitree Go2 API, which fuses onboard LiDAR and joint-encoder data through an Extended Kalman Filter, and the SLAM Toolbox provides pose estimates at 100\,Hz. RGB-D images at 30\,Hz are streamed to \textit{RoboAtlas}, which generates navigation goals $(x,y)$ at 0.07--0.2\,Hz; Nav2 converts these goals into velocity commands via the Unitree \textit{Move} API, which also provides built-in avoidance of moving obstacles.

\textbf{Isaac Sim.} We use Isaac Sim to validate the full closed-loop system under realistic physics and simulated sensors. The same ROS2 stack (SLAM Toolbox, Nav2 SMAC) used on hardware runs against a simulated Go2 with a simulated LiDAR and RGB-D camera, so that the only differences from the hardware setup are the simulated sensor streams and the absence of the Jetson/WiFi link. We use this platform for the CMAB validation in Sec.~\ref{sec:performance-results}, where the robot searches an office environment for a target object.

\textbf{Habitat.} We use the Habitat simulator with HM3D scenes for the large-scale GOAT-Bench evaluation. Here the simulator provides ground-truth agent pose and rendered RGB-D observations directly, so SLAM Toolbox localization is bypassed. \textit{RoboAtlas} consumes the simulator pose and RGB-D render, builds the 3D semantic map and Scene-Dictionary as before, and issues goals that are executed by the Habitat agent's navigation interface rather than Nav2.

We note that most related works benchmark exclusively in Habitat, which abstracts away the principal challenges of physical deployment: real-time inference and planning under a bounded computational budget; localization drift from noisy odometry and real-world SLAM; noisy real-world depth sensing and the imperfect perception built upon it; actuation and footprint constraints; and bounded on-board GPU memory during continuous large-scale operation. Instead, we validate \textit{RoboAtlas} across all three platforms, including physics-based simulation and real hardware.

 \textbf{Edge Deployment (Jetson AGX Thor).} To assess whether \textit{RoboAtlas} is viable without off-board compute, we additionally deploy the complete system onto a single NVIDIA Jetson AGX Thor mounted on the robot, replacing both the desktop RTX~4090 and the WiFi link. In this configuration the Jetson runs the full \textit{RoboAtlas} stack, i.e., \textit{OpenRoboVox} semantic mapping, the 2D pillar and similarity maps, the Scene-Dictionary, and all three experts, together with a locally served Qwen2.5-VL-7B backbone fulfilling both the LLM and VLM roles, concurrently with the ROS2 Active SLAM stack (SLAM Toolbox and Nav2) described above. Perception, mapping, foundation-model inference, goal selection, and motion planning therefore all execute on-board, and the system operates without network connectivity or access to a remote inference API. The decisive enabler is memory capacity rather than raw compute: Thor exposes 128\,GB of LPDDR5X unified across CPU and GPU, and in steady-state operation we measure 43.3\,GiB reserved by the vision-language backbone alongside 3.2\,GiB for the mapping and reasoning process, giving a combined resident footprint of approximately 47\,GiB. This is nearly twice the 24\,GB available on the RTX~4090, so the fully on-board configuration cannot be hosted on the desktop GPU at all, irrespective of throughput. The unified address space also removes the fixed host/device memory partition that produced the out-of-memory failures discussed in Sec.~\ref{methods:gpu_optimization}, allowing the semantic map to grow without contending against a statically sized framebuffer.

\subsection{Performance Results}
\label{sec:performance-results}
\subsubsection{OpenRoboVox}

\begin{figure*}[t]
    \centering

    \includegraphics[width=\textwidth]{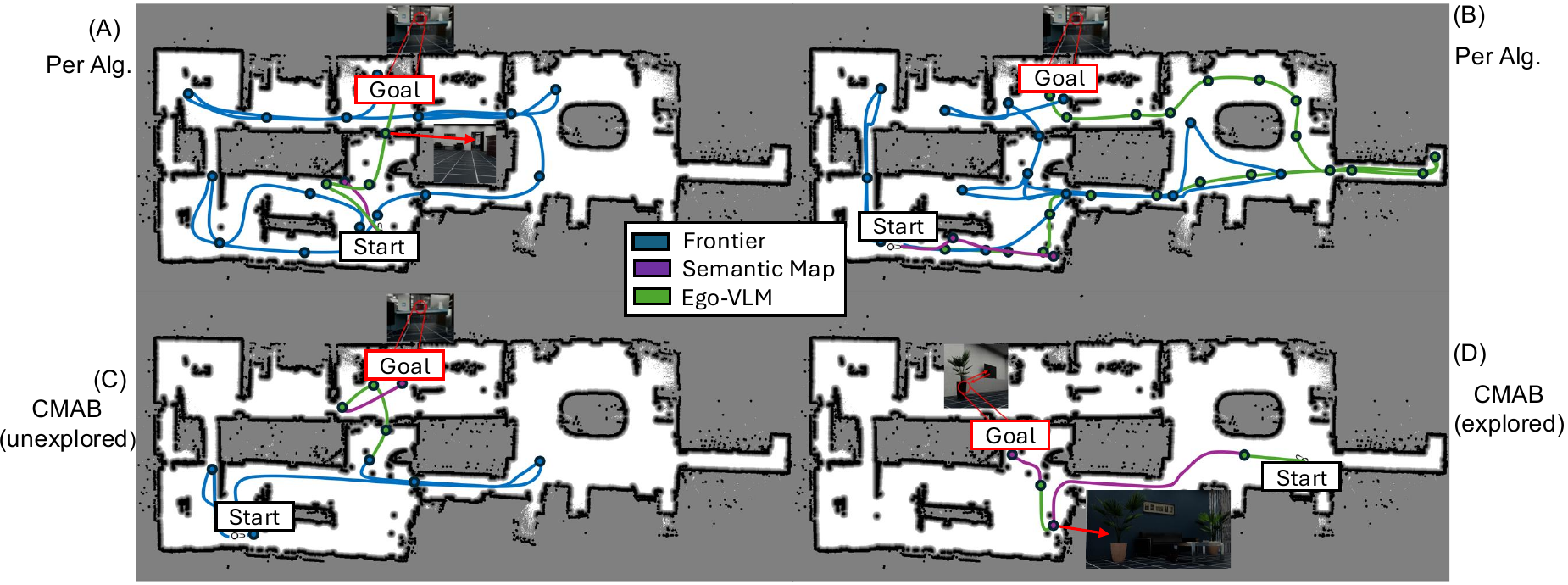}

    \vspace{0.5em}

    \begin{subtable}[t]{0.48\textwidth}
        \centering
        \caption{Navigation performance of four algorithms over 15 trials; map is unexplored. Task success means time the object is found in less than 2000 seconds.}
        \label{tab:nav_results}
        \small
        \resizebox{0.9\linewidth}{!}{%
        \begin{tabular}{lccc}
            \toprule
            \textbf{Algorithm} & \textbf{Success Rate} & \textbf{SPL} & \textbf{Avg. Time (s)} \\
            \midrule
            Semantic Map & 0.0\%   & 0.000 & N/A \\
            Ego-VLM       & 66.7\%  & 0.510 & $964 \pm 90$ \\
            Frontier  & 53.3\%  & 0.386 & $1020 \pm 160$ \\
            CMAB      & 100.0\% & 0.920 & $780 \pm 64$ \\
            \bottomrule
        \end{tabular}%
        }
    \end{subtable}
    \hfill
    \begin{subtable}[t]{0.48\textwidth}
        \centering
        \caption{Navigation performance across four algorithms over 15 trials; map is \emph{pre-explored}. Task success means time the object is found in less than 1000 seconds. Path length and time are reported as mean $\pm$ std over successful trials.}
        \label{tab:nav_results_explored}
        \small
        \resizebox{0.9\linewidth}{!}{%
        \begin{tabular}{lccc}
            \toprule
            \textbf{Algorithm} & \textbf{Success Rate} & \textbf{SPL} & \textbf{Avg. Time (s)} \\
            \midrule
            Semantic Map & 93.3\%  & 0.817 & $610 \pm 44$ \\
            Ego-VLM       & 80.0\%  & 0.638 & $670 \pm 70$ \\
            Frontier  & 33.3\%  & 0.159 & $920 \pm 150$ \\
            CMAB      & 100.0\% & 0.937 & $570 \pm 36$ \\
            \bottomrule
        \end{tabular}%
        }
    \end{subtable}

    \captionsetup{font=small}
    \caption{\textbf{Contextual Multi-Arm Bandit Validation}. Top: Path results based on using only one expert for finding a \textit{large can on the glass table} in (A) - (C) or \textit{green plant next to the tv} in (D). These experts include frontier exploration expert (blue), semantic map expert (purple), or ego-centric VLM expert (green) Bottom: summary statistics over 15 trials for each setting.}
    \label{fig:cmab_results}
\end{figure*}

\begin{figure}[t]
    \centering
\includegraphics[width=.99\columnwidth]{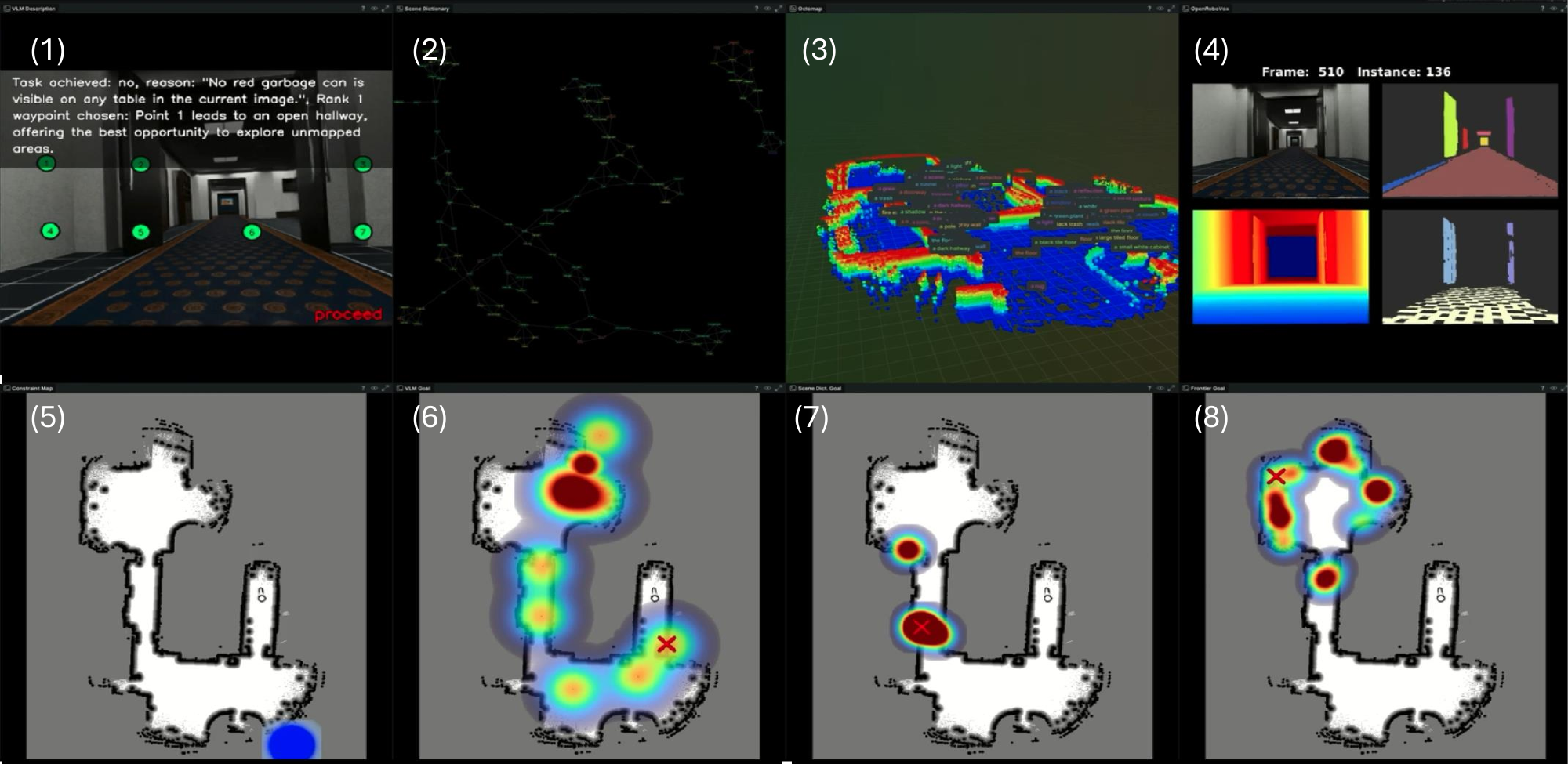} 
    \captionsetup{font=small}
    \caption{\textbf{\textit{RoboAtlas} Demonstration.} Top row: (1) Ego-centric VLM text output, (2) scene dictionary, (3) Octomap visualization with overlaid scene-dictionary captions, (4) OpenRoboVox semantic visualization. Bottom row: (5) constraint map (blue indicates objects to avoid), (6) Ego-centric VLM goal expert (7) semantic map expert, and (8) frontier exploration expert (red X indicates the proposed goal position).}
    \label{fig:roboatlas_demo}
\end{figure}

\begin{figure*}[t]
    \centering
    \includegraphics[width=\textwidth]{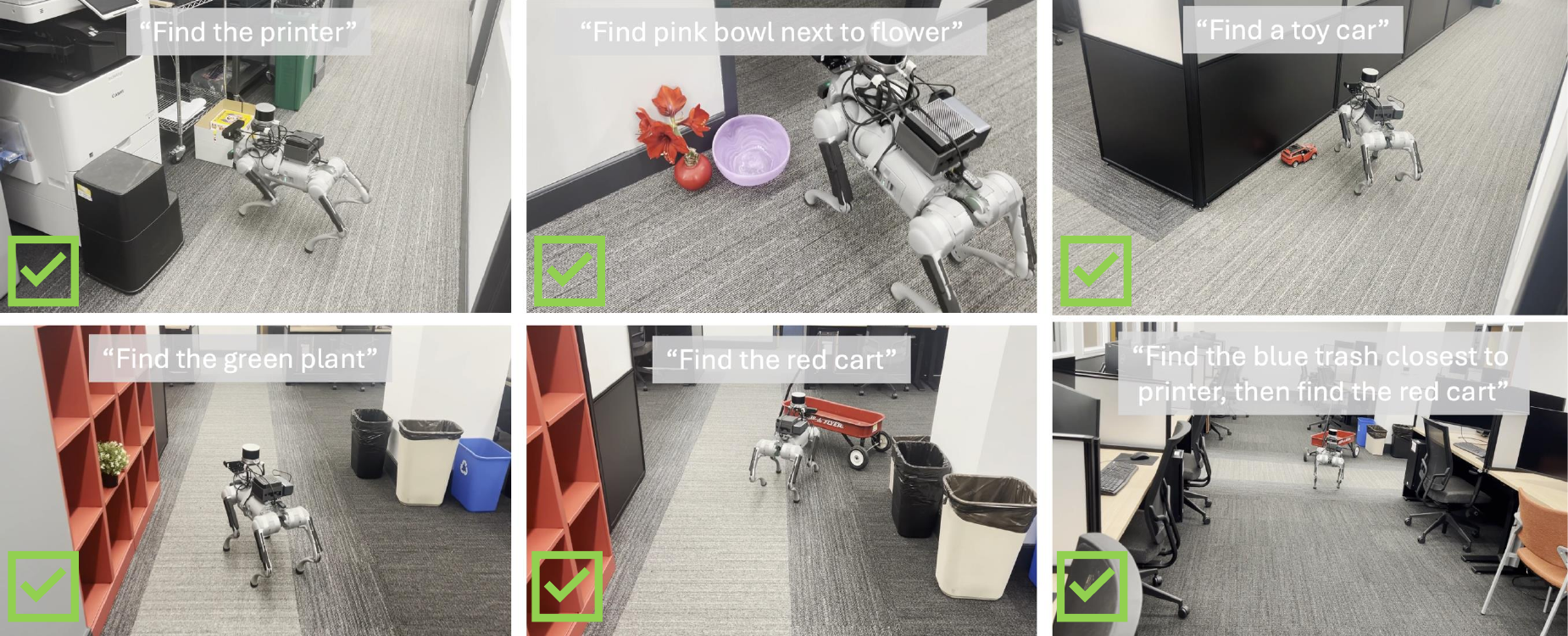}
    \captionsetup{font=small}
    \caption{\textbf{\textit{RoboAtlas.}} Hardware Validation.}
    \label{fig:hardware_demo}
\end{figure*}

\begin{figure*}[!t]
    \centering

    \begin{subfigure}{\textwidth}
        \centering
        \includegraphics[width=\textwidth]{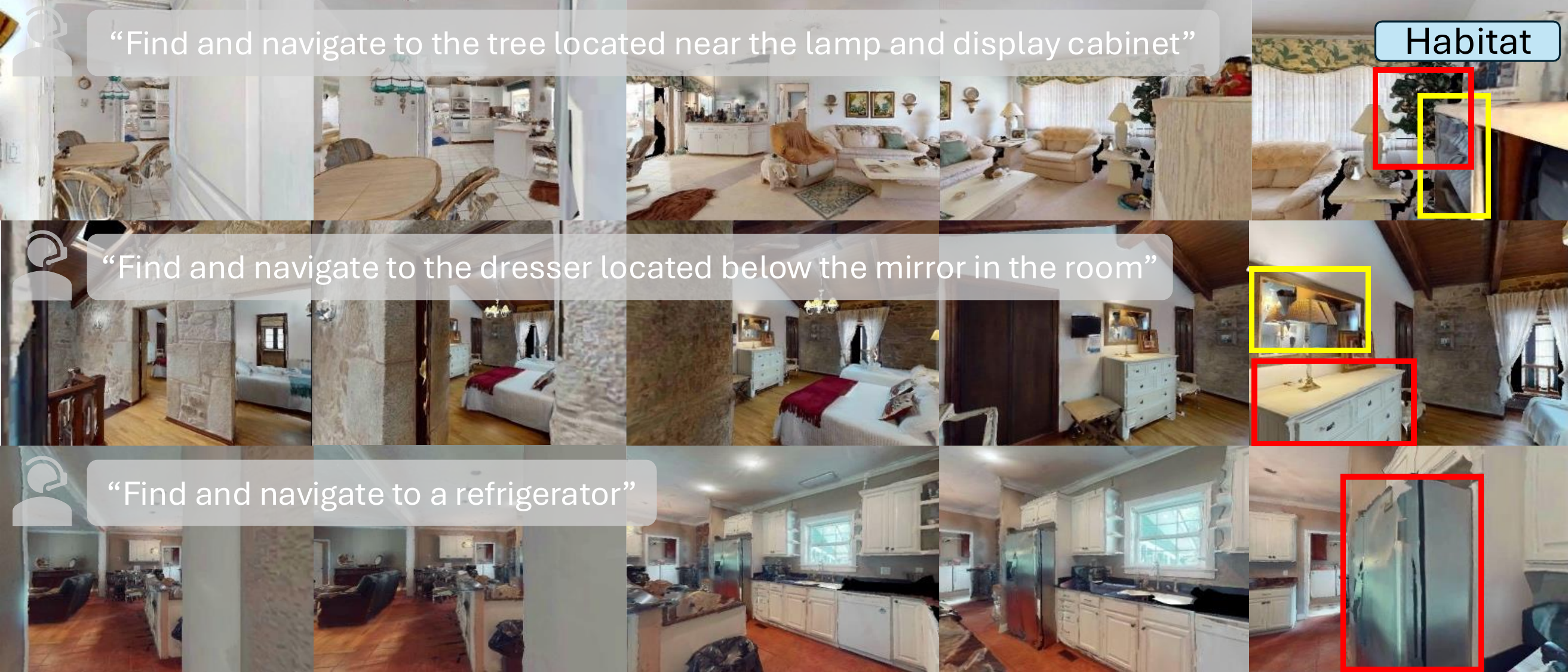}
        \caption{Photo-realistic Habitat simulator validation. Here, we visualize 3 out of 36 val-unseen scenes validated in this study.}
    \end{subfigure}

    \vspace{0.5em}

    \begin{subfigure}{\textwidth}
        \centering
        \includegraphics[width=\textwidth]{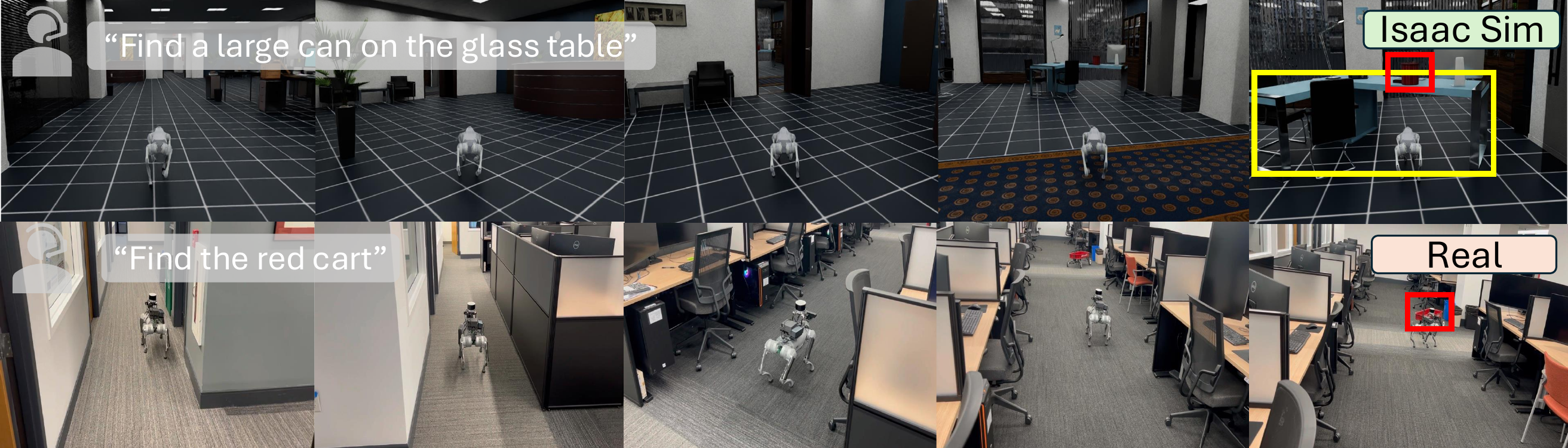}
        \caption{Physics-based Isaac Sim (top) and real-world validation (bottom).}
    \end{subfigure}

    \captionsetup{font=small}
    \caption{\textbf{Cross-Domain Validation}. Red rectangles represent the target object. If neighbor object is specified, they are shown as yellow rectangles.}
    \label{fig:validation_collage}
\end{figure*}

We first validate the \textit{OpenRoboVox} method on hardware and stress test the extend of our 3D semantic mapping capability. In Fig. \ref{fig:octomap_hardware_demo}, we show our capability for creating large 3D semantic maps on hardware, in real-time and on a single-run, for two different floors (left and right columns). As an estimate, we mapped approximately \SI{1803}{\square\meter}, across 2 floors of an office building and identified 29,588 unique instances, which are overlaid on the 3D occupancy grid (bottom row).

\subsubsection{Contextual Multi-Arm Bandits}
\label{cmab_results}

To demonstrate the use of the Contextual Multi-Arm Bandits (CMAB), we present an example scenario in Fig.~\ref{fig:cmab_results} using the office environment from Isaac Sim. See Fig.~\ref{fig:roboatlas_demo} for a visualization of the \textit{RoboAtlas} framework. In (A)–(C), the robot is tasked with finding a \textit{a large can on the glass table} in an environment assumed to be \textit{a priori} unexplored. To highlight the importance of adaptive decision-making, in (A) and (B), where the difference is the robot's starting position, we present 15 trials where the robot is restricted to using a single expert for the entire task: frontier exploration (blue), semantic map (purple), or egocentric VLM (green).

Frontier exploration consistently succeeds but is inefficient, as the robot explores nearly the entire map before locating the target. In contrast, the semantic map expert fails in this setting because it relies on a sufficiently populated scene dictionary; without prior exploration, semantic reasoning is under-constrained. The egocentric VLM exhibits more variable behavior: in some cases it identifies semantically meaningful structures (e.g., doorways) that accelerate discovery, while in others it becomes locally trapped due to limited global context.

CMAB, shown in (C) and (D), dynamically selects among these strategies. In the unexplored case (C), CMAB initially favors frontier exploration to rapidly expand map coverage and collect semantic observations. As the map becomes more complete, it transitions toward egocentric VLM and semantic map experts, enabling semantically guided navigation. In the explored case (D), CMAB leverages semantic reasoning more heavily, combining LLM-based ranking using the semantic expert with VLM-based corrections from the egocentric expert to resolve ambiguity and successfully identify relational targets.

Quantitative results are summarized in Tables (a) and (b) of Fig.~\ref{fig:cmab_results}. We report Success Rate (SR) and Success weighted by Path Length (SPL) at a 1.0\,m threshold (i.e., success means the robot is within 1.0\,m of the object). CMAB achieves a 100\% success rate across all trials while also yielding the shortest path lengths and fastest completion times. In comparison, egocentric VLM achieves a 67\% success rate, while frontier and semantic map experts alone suffer from inefficiencies and failure modes tied to lack of semantic awareness or insufficient exploration.

\vspace{0.5em}
\noindent\textbf{Policy Adaptation Analysis.}
To better understand how CMAB adapts its decision policy, we analyze arm-selection behavior across different exploration stages. We partition decision epochs into low, mid, and high coverage regimes and report the corresponding action distributions in Table~\ref{tab:ucb_actions}. In the low-coverage regime, frontier exploration dominates (63.6\%), reflecting the need for rapid geometric map expansion. In the mid-coverage regime, egocentric VLM becomes more prominent (33.3\%), indicating increased reliance on local semantic cues. In the high-coverage regime, CMAB shifts toward semantic exploitation, selecting semantic map (55.9\%) and egocentric VLM (41.2\%) experts while largely abandoning frontier exploration (2.9\%). This progression demonstrates that CMAB naturally transitions from exploration to semantic reasoning as environmental understanding improves.

\begin{table}[t]
\centering
\captionsetup{font=small}
\caption{UCB-selected action distribution by coverage stage.}
\label{tab:ucb_actions}
\small
\begin{tabular}{c c c c}
\hline
Stage & Frontier & Semantic Map & Ego-VLM \\
\hline
Stage 0 (low)  & 0.636 & 0.303 & 0.061 \\
Stage 1 (mid)  & 0.576 & 0.091 & 0.333 \\
Stage 2 (high) & 0.029 & 0.559 & 0.412 \\
\hline
\end{tabular}
\end{table}

\vspace{0.5em}
\noindent\textbf{Reward Structure Analysis.}
We next examine how the reward function influences this behavior. Table~\ref{tab:reward_by_stage_normalized} reports the mean reward achieved by each arm across coverage regimes, normalized across all arms, for easier comparison. In the low-coverage regime, frontier exploration gained the highest reward ($0.94 \pm 0.36$). In the mid-coverage regime, VLM achieves the highest reward ($0.36 \pm 0.54$), indicating the importance of local visual reasoning when partial semantic context is available. In the high-coverage regime, both semantic map and VLM maintain strong rewards, reflecting the increased utility of semantic reasoning once the scene dictionary is sufficiently populated.



\begin{table}[t]
\centering
\caption{Min--max normalized mean $\pm$ scaled std reward by coverage stage and arm (normalized to $[0,1]$).}
\label{tab:reward_by_stage_normalized}
\renewcommand{\arraystretch}{1.15}
\begin{tabular}{cccc}
\toprule
Stage & Frontier & Semantic Map & Ego-VLM \\
\midrule
Stage 0 (low)  & $0.94 \pm 0.36$ & $0.00 \pm 0.71$ & $0.82 \pm 0.31$ \\
Stage 1 (mid)  & $0.26 \pm 0.53$ & $0.08 \pm 0.55$ & $0.36 \pm 0.54$ \\
Stage 2 (high) & $0.54 \pm 0.96$ & $1.00 \pm 0.70$ & $0.97 \pm 0.57$ \\
\bottomrule
\end{tabular}
\end{table}

To further interpret the reward design, Table~\ref{tab:reward_components} decomposes the total reward into its constituent components. The largest contribution arises from cosine similarity (46\%), followed by VLM confidence (29\%) and task completion (15\%). Coverage rate contributes 13\%, while the backtracking penalty reduces the total reward by approximately 4\%. This distribution indicates that CMAB increasingly favors semantically informative actions as exploration progresses. As the map becomes richer, semantic signals dominate the reward landscape and exert a stronger influence on action selection than purely exploratory incentives.


\begin{table}[t]
\centering
\caption{Reward component statistics over all samples.}
\label{tab:reward_components}
\renewcommand{\arraystretch}{1.15}
\begin{tabular}{lccc}
\toprule
\textbf{Component} & \textbf{Mean} & \textbf{Std} & \textbf{Fraction} \\
\midrule
Coverage rate         & $0.29$  & $0.38$ & $0.13$ \\
Backtracking penalty  & $-0.08$ & $1.27$ & $-0.04$ \\
Cosine similarity     & $1.05$  & $0.71$ & $0.46$ \\
VLM confidence        & $0.66$  & $0.13$ & $0.29$ \\
Task completion       & $0.34$  & $0.38$ & $0.15$ \\
\midrule
Total reward          & $2.26$  & $2.87$ & $1.00$ \\
\bottomrule
\end{tabular}
\end{table}

Overall, these results provide interpretability for CMAB’s performance gains. Rather than relying on a fixed policy, CMAB adapts its behavior according to both the environment state and the reward structure, transitioning from coverage-driven exploration to semantically guided navigation. This adaptive strategy explains why CMAB achieves higher success rates and more efficient trajectories than any single fixed approach.

\begin{table}[t]
\centering
\caption{Hardware validation results. Standoff distance is measured between the robot and the target object at task completion.}
\label{tab:hardware_results}
\renewcommand{\arraystretch}{1.15}
\begin{tabular}{l>{\raggedright\arraybackslash}p{4.5cm}c}
\toprule
\textbf{\#} & \textbf{User Directive} & \textbf{Standoff Distance (m)} \\
\midrule
1 & \textit{``Find the printer''} & 0.20 \\
2 & \textit{``Find pink bowl next to flower''} & 0.05 \\
3 & \textit{``Find a toy car''} & 0.10 \\
4 & \textit{``Find the green plant''} & 0.30 \\
5 & \textit{``Find the red cart''} & 0.15 \\
6 & \textit{``Find the blue trash closest to printer, then find the red cart''} & 0.50 \\
\midrule
\multicolumn{2}{l}{\textbf{Success rate: 100\%}} & \textbf{$0.22 \pm 0.16$} \\
\bottomrule
\end{tabular}
\end{table}

\subsubsection{Hardware Validation}
We further validate the \textit{RoboAtlas} framework on hardware, as shown in Fig.~\ref{fig:hardware_demo}. In all trials, the robot successfully executed the user directives, achieving a 100\% task success rate. Table~\ref{tab:hardware_results} reports the distance error between the robot (measured at the front camera position) and the target object location.

We observe that this error arises from two primary factors. First, small inaccuracies in depth estimation from the onboard camera introduce minor localization discrepancies. More significantly, the robot maintains a safety margin enforced by the obstacle avoidance system, which prevents it from approaching objects too closely. While reducing this safety boundary could decrease the reported distance error, doing so would increase the risk of collision and is therefore undesirable in practice.

\subsection{Baseline Comparisons}

\label{sec:baseline_comparisons}

We evaluate \textit{RoboAtlas} on the GOAT-Bench benchmark~\cite{khanna2024goatbench}, a multimodal lifelong navigation task in which an agent must sequentially navigate to targets specified by object category, language description, or image in unseen environments. Following the evaluation protocol of 3D-Mem~\cite{yang20253dmem} and ReEXplore~\cite{zhang2025reexplore}, we assess the ``Val Unseen'' split using one exploration episode per scene across all 36 scenes, totaling 278 navigation subtasks. As done for our CMAB validation in Sec. \ref{cmab_results}, here a subtask is considered successful if the agent terminates within 1.0\,m of the ground-truth target position defined by GOAT-Bench. 

For this evaluation, \textit{RoboAtlas} operates with the full mixture-of-experts described in Sec.~\ref{methods:main_sec}, including the frontier expert (Sec.~\ref{methods:frontier_selection}, \ref{methods:frontier_generation}), the semantic map expert facilitated by \textit{OpenRoboVox} (Sec.~\ref{methods:openRoboVox-concept}, \ref{methods:semantic_map_expert}), and the egocentric VLM expert (Sec.~\ref{methods:vlm}, \ref{methods:vlm_goal_selection}). Expert selection follows the CMAB policy of Sec.~\ref{sec:cmab}, with the reward structure adapted to the subtask-based protocol of GOAT-Bench as described in Sec.~\ref{sec:goat_reward} of the Appendix of the supplementary material. As detailed there, the bandit operates online from a cold start at the beginning of each scene. Both the context vector $\mathbf{c}_t$ and the supervisory reward are derived purely from agent-perceived state, so the CMAB statistics are updated continuously throughout evaluation.

To interface \textit{RoboAtlas} with GOAT-Bench across all query modalities, we employ a lightweight LLM layer that constructs modality-specific prompts $(\mathcal{Q}_{\mathcal{T}}, \mathcal{Q}_{\mathcal{N}})$ from each GOAT subtask specification and feeds them to the experts described in Sec.~\ref{methods:main_sec}. For description and image subtasks, this layer extracts spatial clue objects from relational phrases (e.g., \textit{``near the table''}, \textit{``next to the dresser''}) and includes them in $\mathcal{Q}_{\mathcal{N}}$ to enrich the Neighbor Similarity computation $\mathcal{N}_{\text{\textrm{sim}}}$ (Eq.~\ref{eq:neighbor_similarity}). Subsequently, the CMAB selects among the frontier, semantic map, and egocentric VLM experts according to the policy in Sec.~\ref{sec:cmab}.

\begin{table}[t]
\centering
\captionsetup{font=small}
\caption{GOAT-Bench results on the ``Val Unseen'' split (278 subtasks, 36 scenes). SR and SPL are reported at the 1.0\,m threshold. Methods marked with $*$ are reported from GOAT-Bench~\cite{khanna2024goatbench}; those with $\dagger$ are evaluated on the same subset.}
\label{tab:goat_benchmark}
\small
\resizebox{\linewidth}{!}{%
\begin{tabular}{lcc}
\toprule
\textbf{Method} & \textbf{SR (\%)} $\uparrow$ & \textbf{SPL (\%)} $\uparrow$ \\
\midrule
\multicolumn{3}{l}{\textit{Traditional Methods}} \\
Modular GOAT$^{*}$ \cite{khanna2024goatbench} & 24.9 & 17.2 \\
Modular CLIP on Wheels$^{*}$ \cite{gadre2022cow} & 16.1 & 10.4 \\
SenseAct-NN Skill Chain$^{*}$ \cite{khanna2024goatbench} & 29.5 & 11.3 \\
SenseAct-NN Monolithic$^{*}$ \cite{khanna2024goatbench} & 12.3 & 6.8 \\
\midrule
\multicolumn{3}{l}{\textit{MLLM-based Exploration}} \\
3D-Mem$^{\dagger}$ (Qwen2.5-VL-7B) \cite{yang20253dmem} & 49.6 & 29.4 \\
ReEXplore$^{\dagger}$ (Qwen2.5-VL-7B) \cite{zhang2025reexplore} & 53.2 & 32.6 \\
HIMM$^{\dagger}$ (Qwen2.5-VL-7B) \cite{li2026himmhumaninspiredlongtermmemory} & 53.9 & 31.7 \\
\addlinespace
Explore-EQA$^{\dagger}$ (GPT-4o) \cite{ren2024explore} & 55.0 & 37.9 \\
CG w/ Frontier Snapshots$^{\dagger}$ (GPT-4o) \cite{gu2024conceptgraphs} & 61.5 & 45.3 \\
3D-Mem w/o memory$^{\dagger}$ (GPT-4o) \cite{yang20253dmem} & 58.6 & 38.5 \\
3D-Mem$^{\dagger}$ (GPT-4o) \cite{yang20253dmem} & 68.9 & 48.9 \\
ReEXplore$^{\dagger}$ (GPT-4o) \cite{zhang2025reexplore} & 59.8 & 42.5 \\
HIMM$^{\dagger}$ (GPT-4o) \cite{li2026himmhumaninspiredlongtermmemory} & 72.8 & \textbf{56.1} \\
\midrule
\multicolumn{3}{l}{\textit{Ours}} \\
\textbf{\textit{RoboAtlas}}$^{\dagger}$ (Qwen2.5-VL-7B) & 88.8 & 53.1 \\
\textbf{\textit{RoboAtlas}}$^{\dagger}$ (GPT-4o) & \textbf{90.6} & 53.4 \\
\bottomrule
\end{tabular}%
}
\end{table}

\begin{table}[t]
\centering
\captionsetup{font=small}
\caption{Per-modality GOAT-Bench results for \textit{RoboAtlas} on the ``Val Unseen'' split, reported at the 1.0\,m success threshold. Results are shown for Qwen2.5-VL-7B and GPT-4o models.}
\label{tab:goat_modality}
\small
\begin{tabular}{l cc cc c}
\toprule
& \multicolumn{2}{c}{\textbf{Qwen2.5-VL-7B}} & \multicolumn{2}{c}{\textbf{GPT-4o}} & \\
\textbf{Modality} & SR (\%) & SPL (\%) & SR (\%) & SPL (\%) & $n$ \\
\midrule
Overall     & 88.8 & 53.1 & 90.6   & 53.4   & 278 \\
Object      & 92.9 & 67.4 & 94.9   & 61.9   &  99 \\
Description & 86.8 & 49.3 & 84.6   & 46.9   &  91 \\
Image       & 86.4 & 40.9 & 92.0   & 50.5   &  88 \\
\bottomrule
\end{tabular}
\end{table}


Table~\ref{tab:goat_benchmark} compares \textit{RoboAtlas} against prior methods on GOAT-Bench. With GPT-4o, \textit{RoboAtlas} attains the highest success rate at 90.6\% SR and 53.4\% SPL at the 1.0\,m threshold, exceeding the strongest prior GPT-4o baseline, HIMM~\cite{li2026himmhumaninspiredlongtermmemory}, by 17.8 percentage points on SR (90.6 vs.\ 72.8), and the next-strongest baseline, 3D-Mem (GPT-4o)~\cite{yang20253dmem}, by 21.7 points (90.6 vs.\ 68.9). 
Notably, even when paired with the smaller Qwen2.5-VL-7B model, \textit{RoboAtlas} achieves 88.8\% SR and 53.1\% SPL. This exceeds all Qwen2.5-VL-7B baselines by a wide margin: ReEXplore~\cite{zhang2025reexplore} by 35.6 points on SR and 20.5 on SPL, 3D-Mem~\cite{yang20253dmem} by 39.2 and 23.7, and HIMM~\cite{li2026himmhumaninspiredlongtermmemory} by 34.9 and 21.4.
Strikingly, \textit{RoboAtlas} (Qwen2.5-VL-7B) outperforms every GPT-4o baseline on both metrics except HIMM. It surpasses 3D-Mem (GPT-4o), for instance, on both SR (88.8 vs.\ 68.9) and SPL (53.1 vs.\ 48.9). Against HIMM, \textit{RoboAtlas} still leads in success rate (88.8 vs.\ 72.8) and trails only in SPL. A 7B backbone competing with GPT-4o-scale baselines suggests the improvements come from our semantic mapping and contextual expert selection rather than backbone capacity.
The per-modality breakdown in Table~\ref{tab:goat_modality} shows that performance is consistent across all three query modalities for both backbones, with object-category subtasks being the strongest (94.9\%/92.9\% SR for GPT-4o/Qwen) and description and image subtasks remaining above 84\% SR.

The one metric in which \textit{RoboAtlas} trails is SPL, where HIMM is marginally higher (56.1 vs.\ 53.4). This reflects a deliberate property of the Active SLAM formulation rather than less effective navigation. \textit{RoboAtlas} maintains exploration pressure throughout each subtask, since the frontier and instance-discovery arms remain eligible to the bandit and the reward favors coverage and discovery, so every navigation rollout doubles as a mapping rollout and yields somewhat longer paths. This is the cost half of an explicit trade: a modest penalty on per-subtask efficiency in exchange for the representational capital that drives the large success margin. The benefit is most pronounced on first-encounter targets, where a direct path to the goal cannot be determined until sufficient portions of the scene have been observed, precisely the case in which prior methods often fail outright. In long-horizon, multi-subtask episodes, the representation constructed during earlier subtasks carries forward to later ones, aligning with the operating regime targeted by GOAT-Bench.

\begin{table}[t]
\centering
\captionsetup{font=small}
\caption{CMAB allocation on GOAT-Bench val-unseen, aggregated over 2{,}005 bandit decisions across 278 subtasks. \emph{Picks} is the number of decisions the CMAB policy assigned to each expert and its share of total decisions; \emph{PL} is the cumulative path length the agent traveled while executing goals from that expert and its share of total travel.}
\label{tab:goat_bandit_telemetry}
\small
\setlength{\tabcolsep}{4pt}
\begin{tabular}{l r r r r}
\toprule
\textbf{Expert} & \textbf{Picks} & \textbf{Picks (\%)} & \textbf{PL (m)} & \textbf{PL (\%)} \\
\midrule
Semantic Map     & 775 & 38.7 & 3360.0 & 46.8 \\
Frontier         & 328 & 16.4 & 1623.1 & 22.6 \\
Egocentric VLM   & 902 & 45.0 & 2201.4 & 30.6 \\
\bottomrule
\end{tabular}
\end{table}

To understand how the CMAB policy distributes work among the three experts (or arms), Table~\ref{tab:goat_bandit_telemetry} reports per-expert pick counts and accumulated path length, aggregated over all 2{,}005 bandit decisions on val-unseen. The egocentric VLM expert receives the largest share of picks (45.0\%) but contributes only 30.6\% of total path length, reflecting its role as a fine-grained corrector that issues many short-range waypoints to refine local positioning. The semantic map expert is selected on 38.7\% of decisions yet accounts for the majority of total path length (46.8\%), consistent with its role as the primary exploitation channel. Once the Scene-Dictionary $\mathcal{S}_t$ contains a high-similarity candidate, this expert dispatches a long-range waypoint toward it. The frontier expert is the least-picked arm (16.4\% of decisions) but contributes a disproportionate 22.6\% of path length, indicating that the bandit reserves it for preliminary exploration, and later for situations where genuine exploration is needed where frontier goals tend to cover substantial distances. Together, the three experts therefore play complementary roles: short-range refinement, long-range semantic exploitation, and occasional long-range exploration. This stage-dependent allocation, learned online from agent-side rewards alone, is the mechanism through which CMAB matches the right expert to the current subtask context.

\section{Discussion and Limitations}
\label{sec:limitations}

The results show that \textit{RoboAtlas} can bridge geometric exploration and semantic reasoning by adapting its decision policy as the environment becomes better understood. Across simulation and hardware experiments, the contextual bandit shifts from coverage-driven behavior toward semantically guided navigation once the scene representation becomes sufficiently informative. The action and reward analyses support this interpretation, showing that semantic signals become increasingly important as map completeness improves.

Some limitations remain for future developments. First, the performance of both LLM-based reasoning and similarity-based retrieval depends strongly on the quality of the underlying 3D semantic map. Missed detections, sparse observations, or incorrect instance associations can propagate into the Scene-Dictionary and produce incorrect goal proposals. This is especially relevant in explored-map settings, where failures can still occur if the target object has not been observed often enough to form a stable semantic representation. These failure modes can be mitigated along two directions. On the algorithmic side, the perception can be strengthened with more accurate open-vocabulary detection and segmentation, and dynamic voxel sizing that can adapt map resolution to object scale, allocating finer voxels to small objects that are otherwise under-represented at a fixed grid resolution. On the hardware side, adding cameras that capture multiple viewpoints would increase observation coverage and reduce missed detections from a limited field of view.

Second, the framework relies on LLM and VLM inference, which introduces latency and variability in response time, particularly for large models such as GPT-4o, which is called through the API. Although the asynchronous system design prevents these delays from blocking the geometric mapping pipeline, the overall performance still depends on network connectivity and remote inference reliability (for GPT-4o model), as the system needs to wait until all actions are available at inference time for the next action to be chosen. This may limit deployment in settings where communication is unreliable or fast global decision making is required, which may be overcome by only using lightweight models (as we have shown with the Qwen2.5-VL-7B model), and consolidating all code onboard, which can be done with more powerful computers such as the Nvidia Jetson Thor.

Third, the Contextual Multi-Arm Bandit (CMAB) depends on a hand-designed reward function that combines geometric and semantic terms. While the reward decomposition in Table~\ref{tab:reward_components} helps explain the observed policy behavior, the relative weighting of these components can influence how aggressively the system explores versus exploits semantic cues. More principled or learned reward formulations may improve generalization across tasks and environments.

A further limitation concerns SPL on GOAT-Bench, which is marginally lower than the best prior method despite our highest success rate. As discussed in Sec.~\ref{sec:baseline_comparisons}, this stems from the Active SLAM formulation: sustained exploration makes navigation rollouts double as mapping rollouts, lowering per-subtask efficiency in exchange for coverage useful to later subtasks.


Overall, these limitations highlight the main challenge in contextual Active SLAM: improving semantic reasoning is not only a modeling problem, but also a systems problem involving perception reliability, inference latency, and robust decision-making under partial observability.
\section{Conclusion}
\label{sec:conclusion}

This paper introduced \textit{RoboAtlas}, a contextual Active SLAM framework that combines geometric exploration and semantic reasoning through a Contextual Multi-Armed Bandit (CMAB). By integrating frontier exploration, LLM-based semantic reasoning, and egocentric VLM cues, the system adapts its navigation strategy to the current state of the environment rather than relying on a fixed exploration policy.

We also presented \textit{OpenRoboVox}, a real-time instance-level semantic mapping framework designed for physical deployment, enabling scalable 3D scene understanding and a language-queryable scene representation. Together, these components allow \textit{RoboAtlas} to operate in both unexplored and semantically complex environments, achieving higher success rates and more efficient behavior than fixed-policy baselines.

The experimental results in simulation, on hardware, and on GOAT-Bench demonstrate that contextual decision-making can improve both robustness and flexibility in Active SLAM. In particular, the results show that adaptive switching between geometric and semantic guidance is effective across different stages of exploration.

Future work will focus on improving robustness to semantic mapping errors, reducing dependence on remote model inference, and extending the framework to longer-horizon real-world deployments. Another promising direction is to compare the CMAB with an end-to-end agentic approach.


\bibliography{bib.bib}
\bibliographystyle{ieeetr}

\appendices
\section{Adapting CMAB for GOAT-Bench}
\label{sec:goat_reward}

The reward formulation in Sec.~\ref{sec:cmab} is designed for a single open-ended directive, where exploration progress and semantic similarity accumulate toward one goal. Instead GOAT-Bench presents a sequence of distinct subtasks per episode, each with its own success criterion. We therefore adapt the bandit pipeline in two ways: (i) the supervisory reward is decomposed into a dense per-turn signal and a sparse per-subtask signal that is propagated to earlier turns via exponential recency weighting; and (ii) the bandit operates strictly on agent-perceived state, so that no benchmark ground truth enters the context vector or the reward. The bandit is initialized cold at the start of each scene and updated continuously throughout evaluation.

\subsection*{Context vector}
At each replanning epoch $t$, the bandit observes a context vector $\mathbf{c}_t \in \mathbb{R}^{18}$ summarizing the agent's current world model. The vector concatenates a bias term, eleven scalar features grouped into three categories, and two one-hot indicators:
\begin{itemize}
    \item \textit{Exploration:} occupancy coverage, instance density, and frontier-set ratio.
    \item \textit{Semantic:} maximum and 90th-percentile cosine similarity between the active task description and Scene-Dictionary embeddings, and the count of above-threshold matches.
    \item \textit{Dynamics:} recent positional displacement, new-instance discovery rate, remaining action budget, fractional task progress, and a repeated-arm indicator over the last three picks.
    \item \textit{Modality and history:} a 3-way one-hot for subtask modality (\textit{object} / \textit{description} / \textit{image}) and a 3-way one-hot for the previously chosen arm.
\end{itemize}
All features are computed from agent-side state alone: the TSDF, the Scene-Dictionary, the ego pose, the navigability and frontier outputs, and the active task string.

\subsection*{Per-turn reward}
At each replanning epoch, a dense per-turn reward $r_t^{\text{turn}}$ combines one-step deltas of four agent-side quantities: the change in occupancy coverage, the change in best target or neighbor similarity to the active task, the new-instance discovery rate, and the magnitude of recent positional displacement. A stuck-indicator penalty is added when the recent-displacement feature falls below a small threshold (suppressed on the first turn after a subtask reset, when no displacement history yet exists). To prevent the bandit from cycling between arms that produce no goal (e.g., a semantic arm whose candidate set is exhausted or an egocentric VLM arm that emits a turning primitive rather than a waypoint), an arm that fails to dispatch a goal receives a small fixed penalty, capped at three empty picks per arm per subtask. An arm that produces empties on a fixed number of consecutive picks is masked from the action set for the remainder of the subtask.

\subsection*{Per-subtask reward}
When subtask $j$ terminates, the bandit issues a sparse supervisory term derived from agent-perceived evidence
\begin{equation}
r^{\text{sub}}_j = w_{\text{succ}} \cdot \mathbf{1}\!\left[\hat{s}_j\right] \cdot \eta_j,
\label{eq:goat_subtask_reward}
\end{equation}
where $\hat{s}_j \in \{0, 1\}$ is an agent-side success proxy that fires only when the agent has both \textit{confidence} and \textit{arrival} evidence for the target. Confidence requires that the maximum Scene-Dictionary similarity to the task description exceeds a threshold $\sigma$, and that a non-trivial number of instances exceed a lower threshold $\tau$. Arrival requires that the recent-displacement feature is below the stuck threshold (the agent has settled) and that at least one high-similarity instance lies within radius $\rho$ of the current ego pose with a label that overlaps the subtask category. The efficiency factor $\eta_j$ equals one for early termination and decays linearly toward a small floor as the action budget is consumed. Subtasks that terminate without firing the proxy contribute zero macro-reward.

The thresholds $\sigma$, $\tau$, and $\rho$ are tuned on a held-out subset of the GOAT-Bench training split, separate from the val-unseen scenes used for evaluation. Similarities are computed by the agent's SBERT encoder against captions produced online by the perception stack, and proximity is evaluated between two agent-frame quantities (the ego pose and the Scene-Dictionary instance position). The benchmark's val-unseen success indicator and viewpoint set are read only by an outcome-only checker used for evaluation reporting and are not visible to the bandit.

\subsection*{Recency-weighted credit assignment}
Let $\{(a_{t_i}, \mathbf{c}_{t_i}, r_{t_i}^{\text{turn}})\}_{i=1}^{n}$ denote the sequence of arms, contexts, and per-turn rewards recorded during subtask $j$. Once $r^{\text{sub}}_j$ is computed at termination, it is distributed across the $n$ turns of the subtask through an exponentially decaying weight
\begin{equation}
r_{t_i} = r_{t_i}^{\text{turn}} + \gamma^{\,n-1-i}\, r^{\text{sub}}_j, \qquad \gamma \in (0, 1).
\label{eq:goat_credit_assignment}
\end{equation}
The arm chosen on the final turn of a successful subtask receives nearly the full macro-reward, while earlier turns receive geometrically less. This biases the policy toward arms that are proximally responsible for the outcome while still crediting earlier choices (e.g., frontier expansion that first uncovered the target instance). The combined $r_{t_i}$ drives the standard LinUCB update of $\mathbf{A}_{a_{t_i}}$ and $\mathbf{b}_{a_{t_i}}$ in Algorithm~\ref{alg:linucb}.

\subsection*{Online operation}
Because both $\mathbf{c}_t$ and $r_{t_i}$ are derived purely from agent-perceived state, the bandit can be updated online on the evaluation set without violating the benchmark protocol. We initialize the per-arm sufficient statistics $\{\mathbf{A}_a = \lambda \mathbf{I},\ \mathbf{b}_a = \mathbf{0}\}$ at the start of each scene and update them continuously as the controller proceeds. This online update aligns the bandit's adaptation timescale with the subtask granularity at which behavioral evidence accumulates, and avoids a train/freeze split that would be required if the reward depended on ground-truth viewpoints or geodesics.

\end{document}